\documentclass[11pt]{article}

\usepackage[final]{acl}

\usepackage{times}
\usepackage{latexsym}

\usepackage[T1]{fontenc}

\usepackage[utf8]{inputenc}

\usepackage{microtype}

\usepackage{inconsolata}

\usepackage{float}
\usepackage{graphicx}
\usepackage{booktabs}
\usepackage{makecell}
\usepackage[autostyle=true]{csquotes}   
\usepackage{multirow}
\usepackage{longtable}
\usepackage{amsmath, amssymb}
\usepackage{tabularx}
\usepackage{tcolorbox}
\tcbuselibrary{breakable, skins}
\usepackage{xcolor}
\usepackage{caption}
\newtcolorbox{promptbox}[1]{
    colback=gray!5!white,
    colframe=gray!75!black,
    fonttitle=\bfseries,
    title=#1,
    breakable,
    boxrule=0.5pt,
    left=1em, right=1em, top=1em, bottom=1em
}
\newcommand\blfootnote[1]{%
  \begingroup
  \renewcommand\thefootnote{}\footnote{#1}%
  \addtocounter{footnote}{-1}%
  \endgroup
}
\title{What Limits Us? Analyzing Self-Reported Limitations in NLP Research}

\author{
    Tawan Thaeprasit$^\dagger$, Peeranuth Kehasukcharoen$^\dagger$, Ding Wang$^\ddagger$,\\
    \bf{Remi Denton$^\ddagger$, Peerapon Vateekul$^{\dagger*}$, Piyawat Lertvittayakumjorn$^\ddagger$} \\
    $^\dagger$Department of Computer Engineering, Faculty of Engineering,\\
    Chulalongkorn University, Thailand \\
    $^\ddagger$Google Research \\
    \texttt{tawanth.official@gmail.com, \ 6030416021@alumni.chula.ac.th} \\
    \texttt{peerapon.v@chula.ac.th, \{drdw,dentone,piyawat\}@google.com}
}

\begin{document}
\maketitle

\begin{abstract}
Since late 2022, a Limitations section has become mandatory at many top-tier NLP conferences. 
The growing number of accepted papers at these venues has resulted in a vast corpus of self-reported limitations that cannot all be manually reviewed, yet remains systematically unanalyzed.
Therefore, in this paper, we conduct a large-scale analysis of the Limitations sections from ACL and EMNLP papers published between 2020 and 2025 to understand what researchers disclose about their own work.
To do so, we implement a novel human-AI framework for iterative hybrid qualitative coding.
This framework enables us to investigate trends in self-reported limitations over time, their correlations with specific paper attributes, and the writing patterns that recur around these disclosures.
Our findings offer a critical reflection on the diverse reported challenges as well as the self-reporting practices of researchers in the NLP community.
\end{abstract}

\section{Introduction}
Reporting research limitations is fundamental to scientific transparency, as it declares factors that might undermine the validity of research claims and provides suggestions for readers who want to use or build upon the work \cite{olteanu2026rigor}.\blfootnote{$*$ Corresponding author}
In the NLP community, the Limitations section was optional until EMNLP 2022 and ACL 2023 made it mandatory for all accepted papers. This policy was subsequently adopted by ACL Rolling Review (ARR) in December 2023\footnote{\url{https://aclrollingreview.org/cfp}}, ensuring that recent papers across all main *ACL conferences include a self-reported Limitations section. Consequently, thousands of these sections are now scattered across papers hosted on the ACL Anthology\footnote{\url{https://aclanthology.org/}}. 
This data availability has enabled several studies involving the Limitations sections of NLP research \cite{al-azher-etal-2025-bagels,10.1007/978-3-031-70344-7_7}. 

Focusing on limitation analysis, \citet{zhou-etal-2025-culture} manually examined the Limitations sections of 57 cultural NLP papers from 2022 to 2024 to understand recurring challenges in the subfield. Meanwhile, \citet{10.1145/3677389.3702605} applied topic modeling to papers from ACL 2023 and its workshops to discover common limitation topics.
However, given the massive volume of papers nowadays, manual analysis is infeasible at scale, while topic modeling tends to bias toward dominant topics and overlook minor yet significant ones, especially newly emerging topics. 
Furthermore, none of the existing work has shed light on how self-reported limitations evolve given the field's rapid progress and the publication policy change.

In this paper, we develop a human-AI framework to conduct large-scale hybrid (deductive and inductive) qualitative coding \cite{doi:10.1177/160940690600500107} on the Limitations sections of ACL and EMNLP papers published between 2020 and 2025. Within this framework, a Large Language Model (LLM) is used to apply existing codes to the text, suggest new codes when needed, and recommend codebook modifications based on emerging new code clusters. Still, humans are responsible for initial codebook development and actual codebook modifications to ensure interpretability of the output. Following the full-scale annotation, the resulting qualitative codes were aggregated into code distributions across the dataset, providing empirical evidence to answer our three core research questions:

\begin{itemize}
\item \textbf{RQ1: Temporal Analysis} -- How have the contents of self-reported limitations evolved over time? (Section \ref{sec:rq1})
\item \textbf{RQ2: Correlations with Paper Attributes} -- How do specific limitation codes correlate with certain paper attributes (i.e., research areas, paper formats, and author affiliations)? (Section \ref{sec:rq2})
\item \textbf{RQ3: Discursive Patterns} -- Are there recurring textual patterns in self-reported limitations that could suggest potential reporting strategies? (Section \ref{sec:rq3})
\end{itemize}

Overall, our contribution is threefold.
First, we propose a novel human-AI framework for iterative hybrid qualitative coding, enabling scalable and rigorous analysis of research limitations.
Second, we release a large-scale dataset of 16,067 Limitations sections extracted from ACL and EMNLP papers (2020--2025). This release includes LLM-generated annotations and a high-quality human-annotated set for evaluation.\footnote{The dataset is released at \url{https://github.com/Sundione/nlp-self-reported-limitations}.} 
Third, we present empirical findings for our three research questions, providing deep insights into trends and practices of self-reported limitations in NLP research.

\section{Background and Related Work}

\paragraph{Analyzing Limitations Sections.} 
Due to the rich insights contained in Limitations sections, researchers across various disciplines have analyzed these sections to understand common challenges within their respective fields \cite{rodriguez2024self,10.1371/journal.pone.0305970,alvarez2021sample,stockli2023reporting,sanders2023computing,theofanidis2018limitations,brutus2013self}. Most of these studies were conducted manually or through keyword analysis. 
Because reporting limitations was not standard practice in NLP until late 2022, literature analyzing these sections within the NLP community remains relatively scarce. 
Nonetheless, following recent conference policy changes, emerging work has begun focusing on limitation extraction and generation \cite{al-azher-etal-2025-bagels,10.1007/978-3-031-70344-7_7} as well as content analysis \cite{zhou-etal-2025-culture}. 
Among existing studies, our work is closest to \citet{10.1145/3677389.3702605}, who applied topic modeling to Limitations sections and used LLM-based summarization to produce topic summaries. 
However, the inherent nature of topic modeling suppresses the detection of minor, emerging topics, which are crucial for temporal analysis. 
Our work adopts a well-established approach in qualitative research to overcome this limitation.

\paragraph{Coding for Qualitative Analysis.}
In qualitative analysis, coding is the process of assigning concise, descriptive labels to specific segments of text so as to understand recurring concepts in textual data \cite{saldana2021coding}.
Researchers generally approach coding through two distinct lenses. 
First, deductive coding relies on a predefined list of codes and their definitions (compiled within a codebook) to analyze the text. 
Conversely, with inductive coding, researchers construct codes bottom-up based on their interpretations of the data.
As a middle ground, a hybrid approach starts with an initial codebook but remains open to generating new, inductive codes for text segments that do not fit any existing codes \cite{doi:10.1177/160940690600500107}.
In this study, we adopt the hybrid approach to analyze Limitations sections, leveraging an existing taxonomy of limitation types \cite{xu-etal-2025-llms-identify} while leaving room for novel patterns to emerge from the data.

To analyze large textual datasets, researchers have increasingly explored AI-assisted tools for qualitative analysis. Early work primarily utilized topic modeling techniques, such as Latent Dirichlet Allocation \cite{10.5555/944919.944937} and BERTopic \citep{grootendorst2022bertopicneuraltopicmodeling}. However, these methods often provide limited support for interpreting the underlying topics. More recently, studies have investigated the use of LLMs to facilitate coding, spanning both deductive \citep{10.1145/3581754.3584136, chew2023llmassistedcontentanalysisusing} and inductive (data-driven) \citep{dai2023llmintheloopleveraginglargelanguage, parfenova-etal-2025-text, zhong-etal-2025-hicode, 10.1145/3801096}  approaches. 
Closest to our work, \citet{wiebe2025qualitative} proposed using an LLM with humans in the loop for hybrid coding and thematic analysis. However, a key distinction is that our approach allows for iterative updates to the codebook, which are essential for temporal analysis. Additionally, our framework is designed in a modular fashion, enabling researchers to easily extend or modify individual modules to suit their specific needs.

\section{Methodology}
\label{sec:methodology}

This section discusses the scope of our study, the content extraction process, and our human-AI framework for hybrid qualitative coding. 

\subsection{Scope of the Study}
We scoped our analysis to ACL and EMNLP papers (Long, Short, and Findings) published between 2020 and 2025. 
We focused exclusively on self-reported limitations within dedicated Limitations sections.
Consequently, any limitations discussed in other parts of the text, or omitted from the papers, were not included in this study.

\subsection{Content Extraction}

To prepare the textual data for analysis, we extracted the Limitations sections along with relevant paper attributes needed to answer RQ2.

\paragraph{Limitations Sections.}
We extracted the limitations from two sources.
For papers prior to EMNLP 2022, their raw texts are provided by the ACL-OCL dataset \cite{rohatgi-etal-2023-acl} from which we collected the Limitations sections. 
For the remaining papers, we developed an extraction pipeline 
using Docling\footnote{\url{https://github.com/docling-project/docling}} to parse the PDF files from ACL Anthology, followed by a regular expression module to extract the target sections.
To validate our pipeline, we cross-checked its output against the ACL-OCL data for ACL 2022 (Short) and found our tool produced consistent results.

\paragraph{Paper Attributes.}
\label{paper_attibutes}
The metadata extracted from each paper consists of research areas and author affiliations. To identify research areas (e.g., Information Extraction, Machine Translation, NLP Applications, etc.), we used information from conference programs available online for ACL 2022 and EMNLP 2023 to 2025. 
For affiliations, we extracted them directly from the raw PDF file provided on the ACL Anthology using a custom extraction pipeline. Further details are provided in Appendix \mbox{\ref{appendix:affiliation_ext}}.

\subsection{Iterative Hybrid Qualitative Coding}

\begin{figure*}[t]
    \centering
    \includegraphics[width=0.95\linewidth]{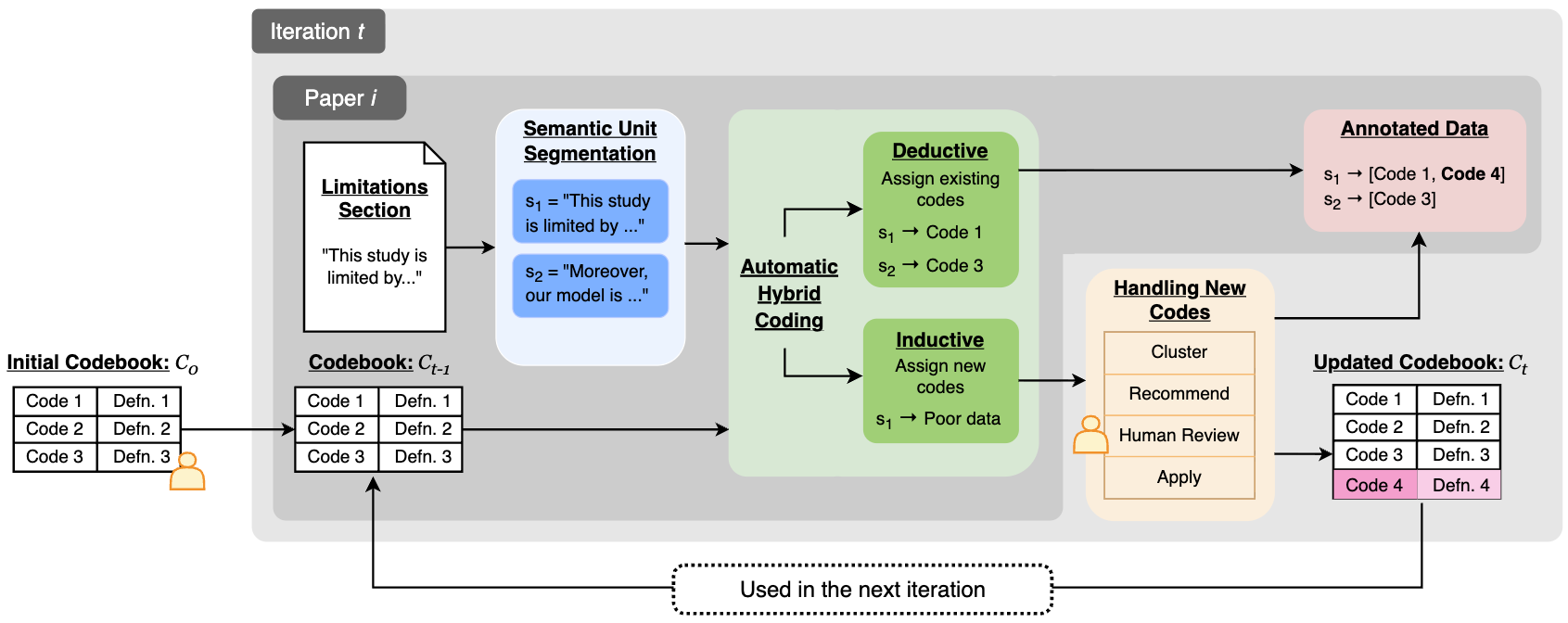}
    \caption{An overview of our iterative hybrid coding framework. 
    The human icons illustrate where humans perform tasks in this framework. ``Defn.'' stands for the definition of the code.
    More details can be found in Section~\ref{limitation_coding}.
    }
    \label{fig:pipeline_flow} 
\end{figure*}

Our hybrid coding framework begins with the construction of an initial codebook, followed by an iterative coding process.
In each iteration, we code the Limitations sections from specific years and update the codebook to incorporate newly found codes.
In this study, we conducted four iterations: the first combined 2020–2022 data (grouped due to a lower volume of Limitations sections), followed by separate iterations for 2023, 2024, and 2025.

\paragraph{Initial Codebook Construction.}
A codebook is a structured set of categories and definitions used to classify text. 
We built our initial codebook by combining data-driven discovery with an established taxonomy.
First, we conducted a pilot run by using an LLM to inductively code 50 Limitations sections from ACL 2024 extracted by BAGELS \citep{al-azher-etal-2025-bagels}. After that, we manually aligned these empirical codes with the taxonomy of limitations in AI research proposed by \citet{xu-etal-2025-llms-identify} and expanded the taxonomy to accommodate novel topics observed in the pilot data. 
To support the analysis of discursive patterns in RQ3, we also created codes for non-limitation content such as future work, method strengths, and conducted mitigation, as seen in the pilot data.   
The resulting codebook, consisting of 18 limitation and 7 non-limitation codes, served as our initial codebook $\mathcal{C}_0$ for the iterative coding process, as displayed in Figure~\ref{fig:pipeline_flow}.

\paragraph{Automatic Hybrid Coding.} \label{limitation_coding}
For each iteration, we used LLMs to code the Limitations section of each paper. Given a Limitations section $L_i$ of a paper $i$, we first segmented the section into a sequence of semantic units $S_i$. Formally,  
\[S_i = \langle s_{i1}, \ldots, s_{in} \rangle = M_{\text{seg}}(L_i)\]
where $s_{ij}$ is a semantic unit in $L_i$ and $M_{\text{seg}}$ is a segmenter model, which is, in our case, Gemini 2.5 Flash \citep{comanici2025gemini25pushingfrontier} running a segmenter prompt. 
Next, we formatted these segmented units with XML tags and grouped them into
batches for coding. A \emph{batch} is the set of semantic units passed to
$M_{\text{code}}$ in a single call, adopted for efficient inference.
We used Gemini 3.1 Pro\footnote{\url{https://ai.google.dev/gemini-api/docs/gemini-3}} as $M_{\text{code}}$ where our coding prompt (in Appendix~\ref{sec:coding-prompt}) employed two prompt engineering strategies. 
First, we provided few-shot examples ($\mathcal{E}$) of semantic units coded by the authors for in-context learning. 
Second, we applied Chain-of-Thought (CoT) prompting, requiring the model to articulate its reasoning before classification.
So, for iteration $t$, 
\[\mathcal{I}_{t,b} = M_{\text{code}}(\mathcal{B}_{t,b},\, \mathcal{C}_{t-1},\, \mathcal{E})\]
where $\mathcal{B}_{t,b}$ is the $b$-th batch of semantic units in iteration $t$, 
$\mathcal{C}_{t-1}$ is a codebook from the previous iteration, 
$\mathcal{E}$ is a set of few-shot examples, 
and $\mathcal{I}_{t,b}$ is the resulting set of coded units in this batch $b$.
Each member in $\mathcal{I}_{t,b}$ is a tuple of semantic unit and an assigned code where one semantic unit can appear multiple times in $\mathcal{I}_{t,b}$. 
At the end, we obtained $\mathcal{I}_t = \bigcup_{b} \mathcal{I}_{t,b}$ as the set of coded units from all the batches in iteration $t$. 

To support hybrid coding, the prompt of $M_{\text{code}}$ allowed the model to assign new codes that are not in $\mathcal{C}_{t-1}$ to specific semantic units if necessary. 
Hence, $\mathcal{I}_t$ can be partitioned into $\mathcal{I}^E_t$ and $\mathcal{I}^N_t$, representing the sets of existing-code assignments and new-code assignments, respectively. 
Next, we explain how we handled the new codes in $\mathcal{I}^N_t$.

\paragraph{Handling New Codes.}
\label{new_code_handling}

Practically, the same new code could be phrased differently by $M_{\text{code}}$ in different batches. Also, some of the new codes could be noises or were not different enough from an existing code.
We therefore refrained from immediately assigning new codes in $\mathcal{I}^N_t$ or updating the codebook without human review. However, given the large volume of papers and new codes in each iteration, manually reviewing every single code assignment in $\mathcal{I}^N_t$ was unfeasible. So, we handled $\mathcal{I}^N_t$ in four steps.
First, an LLM consolidator ($M_{\text{consolidate}}$) grouped semantically similar new codes into clusters $m_i$ with suggested cluster names and definitions.
\[\mathcal{M}_t = \{m_1, \ldots, m_k\} = M_{\text{consolidate}}(\mathcal{I}^N_t)\]
Second, a recommendation module ($M_{\text{rec}}$) evaluated $\mathcal{M}_t$ against the current codebook $\mathcal{C}_{t-1}$ and suggested an action $a_i$ (which could be $\textsc{add}$, $\textsc{merge}$, $\textsc{expand}$, or $\textsc{reject}$) for each $m_i$. 
\[\mathcal{A}_t = \{a_1, \ldots, a_k\} = M_{\text{rec}}(\mathcal{M}_t,\, \mathcal{C}_{t-1})\]
The explanations of each possible action can be found in Table \ref{tab:actions}.
Note that we used Gemini 3.1 Pro for both $M_{\text{consolidate}}$ and $M_{\text{rec}}$.
Third, as the human-in-the-loop researchers, we reviewed $\mathcal{A}_t$ to confirm or modify each action $a_i$ as appropriate, resulting in the set of human-verified actions $\mathcal{A}'_t$.
Finally, we applied $\mathcal{A}'_t$ to update the codebook to $\mathcal{C}_{t}$ and adjusted the code assignments in $\mathcal{I}^N_t$ according to Table \ref{tab:actions}.
This concludes an iteration of hybrid coding. We then began the next iteration of automatic coding with the updated codebook $\mathcal{C}_{t}$.

\begin{table}[t]
\centering
\footnotesize
\setlength{\tabcolsep}{4pt} 
\renewcommand{\arraystretch}{1} 

\begin{tabularx}{0.98\columnwidth}{l >{\hsize=0.37\hsize}X >{\hsize=0.39\hsize}X}
\toprule
\textbf{Action} & \textbf{Codebook Update} & \textbf{$\mathcal{I}^N_t$ Update} \\ 
\midrule
\textsc{Add} & Add a new code $c$ to the codebook for this cluster & 
Change every new code in this cluster to $c$ \\ 
\midrule
\textsc{Expand} & Expand the definition of an existing code $c$ to cover this cluster & Change every new code in this cluster to $c$ \\ 
\midrule
\textsc{Merge} & Use an existing code or a new code $c$ for this cluster (without definition update) & Change every new code in this cluster to $c$ \\ 
\midrule
\textsc{Reject} & Drop this cluster & Remove every new code in this cluster \\ 
\bottomrule
\end{tabularx}
\caption{Possible actions for each new code cluster and the corresponding codebook updates and new-code assignment $\mathcal{I}^N_t$ updates.}
\label{tab:actions}
\end{table}

\subsection{Evaluation of the Hybrid Coding Step}
\begin{table*}[t]
\centering
\footnotesize
\setlength{\tabcolsep}{3pt}
\begin{tabular}{l r r r r r r r r r r}
\toprule
& \multicolumn{4}{c}{\textbf{Deductive}} & \multicolumn{3}{c}{\textbf{Inductive - Quality}} & \multicolumn{3}{c}{\textbf{Inductive - Sensitivity}} \\ \cmidrule(lr){2-5} \cmidrule(l){6-8} \cmidrule(l){9-11}
\textbf{Setting} & \multicolumn{1}{r}{\textbf{Prec. $\uparrow$}} & \multicolumn{1}{r}{\textbf{Recall $\uparrow$}} & \multicolumn{1}{r}{\textbf{F1 $\uparrow$}} & \multicolumn{1}{r}{\textbf{$\alpha$}} & \multicolumn{1}{r}{\textbf{Correct $\uparrow$}} & \multicolumn{1}{r}{\textbf{Partial}} & \multicolumn{1}{r}{\textbf{Wrong $\downarrow$}} & \multicolumn{1}{r}{\textbf{New $\uparrow$}} & \multicolumn{1}{r}{\textbf{Existing}} & \multicolumn{1}{r}{\textbf{Missed $\downarrow$}}\\
\midrule
\citet{10.1145/3581754.3584136} & 0.744 & 0.792 & 0.751 & 0.627 & N/A & N/A & N/A & N/A & N/A & N/A\\
Ours (Defn. only) & 0.757 & 0.760 & 0.738 & 0.623 & 87.5\% & 0.0\% & 12.5\% & 1.4\% & 78.9\% & 19.7\%\\
Ours (Few-Shot only) & 0.786 & 0.808 & 0.776 & 0.635 & 82.0\% & 11.5\% & 6.6\% & 32.4\% & 56.3\% & 11.3\%\\
Ours (Defn. + Few-Shot) & 0.827 & 0.811 & \textbf{0.801} & \textbf{0.647} & \textbf{91.5\%} & 3.4\% & 5.1\% & \textbf{60.6\%} & 35.2\% & 4.2\%\\
\bottomrule

\end{tabular}
\caption{
Performance of different prompting approaches. ``Prec.'' and ``Defn.'' stand for precision and code definitions, respectively. $\alpha$ represents the inter-annotator agreement (Krippendorff's $\alpha$) after treating the LLM as another annotator. Note that the baseline $\alpha$ from human annotators only is 0.644. N/A means the method cannot produce inductive codes by design. Bold numbers highlight key metrics we used for method selection.
}
\label{tab:pipeline_eval}
\end{table*}

Because the semantic unit segmentation step (by $M_{\text{seg}}$) is straightforward and the process for handling new codes (using $M_{\text{consolidate}}$ and $M_{\text{rec}}$) is reviewed by humans, we focus our framework evaluation on the hybrid coding step performed by $M_{\text{code}}$.

\paragraph{Dataset.} We sampled the Limitations sections from 150 NLP papers, 30 of which were labeled by three authors of this paper to calibrate our understanding of the codebook, while the remaining 120 were labeled by two authors. 
Each annotator independently used the initial codebook $\mathcal{C}_0$ for coding and proposed new labels as necessary. 
The inter-annotator agreement (Krippendorff's $\alpha$) of deductive code assignments was 0.644.
Disagreements in both deductive and inductive coding were discussed until a consensus was reached among all annotators assigned to that paper. 

\paragraph{Metrics.} 
We divided the evaluation into two parts. 
For the deductive part, which resembles a multi-label classification task, we computed precision, recall, and F1 scores against the human consensus codes. 
Additionally, we treated the LLM as a supplementary annotator and recomputed Krippendorff's $\alpha$ to observe how LLM-generated labels impacted the baseline inter-annotator agreement.
For the inductive part, we manually assessed the quality of each new code from the LLM, classifying whether it was acceptable, required a minor label adjustment, or was a false positive.
Furthermore, to evaluate the LLM's sensitivity to new codes, we checked whether the LLM detected each new code from human consensus as a new code, assigned an acceptable existing code, or missed it completely.

\paragraph{Comparison.} We compared three variants of our $M_{\text{code}}$, which differ in how they describe the codes within the codebook. These variants evaluate the use of: (1) code definitions only, (2) few-shot examples only, and (3) both code definitions and few-shot examples in the prompt. 
Furthermore, we compared our approaches against a baseline from \citet{10.1145/3581754.3584136}, which utilizes a zero-shot deductive coding prompt that includes the code definitions only. All tested approaches were executed using the same underlying model, i.e., Gemini 3.1 Pro.

\paragraph{Results.}
As Table \ref{tab:pipeline_eval} shows, the deductive baseline \cite{10.1145/3581754.3584136} achieved a competitive F1-score (0.751) for existing codes; however, it failed to generate meaningful new codes. This highlights a fundamental limitation of strictly deductive prompts.
Meanwhile, our approach with only code definitions slightly degraded the deductive metrics compared to the baseline (F1 = 0.738).
In contrast, providing only few-shot examples improved the deductive (F1 = 0.776). Ultimately, our combined ``Defn. + Few-Shot’' setting achieved the highest classification performance (F1 = 0.801). 
Moreover, while the other settings decreased the inter-annotator agreement ($\alpha$), the combined ``Defn. + Few-Shot’' slightly improved $\alpha$ from the baseline of 0.644 to 0.647. This shows that it acted as a valid automated coder, diverging from humans only as much as the humans disagreed with one another.

Regarding inductive coding, we can see from Table~\mbox{\ref{tab:pipeline_eval}} that the ``Defn. + Few-Shot’' setting also achieved the highest metrics for both the new code quality (91.5\%) and sensitivity (60.6\%).
Although the latter reveals some room for improvement, the method rarely ignored novel evidence completely (4.2\%) but tends to conservatively mapped the evidence to acceptable existing codes (35.2\%), making it analytically safe for our large-scale analysis. Hence, we used this combined configuration in our full-scale run.

\section{Results}
In total, we analyzed Limitations sections from 16,067 papers (7,052 ACL and 9,015 EMNLP papers) across four iterations, resulting in the final codebook presented in Appendix~\ref{app:codebook}. 
In this section, we discuss the rate of limitation reporting alongside our findings for the three RQs.

\begin{table}[t]
    \centering
    \footnotesize
    \renewcommand{\arraystretch}{1.2}
    \setlength{\tabcolsep}{3.2pt}
    \begin{tabular}{lcccc}
        \toprule
        \textbf{Venue} & \makecell[c]{\textbf{Total}\\\textbf{Papers}} & \makecell[c]{\textbf{(1)}\\\textbf{Implicitly}\\\textbf{Report$^*$}}& \makecell[c]{\textbf{(2)}\\\textbf{Truly}\\\textbf{Missing$^*$}}& \makecell[c]{\textbf{(3)}\\\textbf{Pipeline}\\\textbf{Failure}}\\
        \midrule
        EMNLP 2022 & 1,376 & 20 & 10 & 0\\
        ACL 2023 & 1,976 & 13 & 9 & 13\\
        EMNLP 2023 & 2,106 & 5  & 11 & 3\\
        ACL 2024 &  1,915 & 4  & 7 & 6\\
        EMNLP 2024 & 2,271  & 2  & 1 & 4\\
        ACL 2025 & 3,086 & 6  & 7 & 7\\
        EMNLP 2025 & 3,214  & 0  & 0 & 8\\
        \midrule
        \textbf{Total} & \textbf{15,944} & \textbf{50} & \textbf{45} & \textbf{41} \\
        \bottomrule
    \end{tabular}
    \caption{Number of accepted papers across venues post-mandate (2022--2025), along with a breakdown of papers flagged by our extraction pipeline as missing a required Limitations section. It distinguishes (1) papers implicitly reporting limitations in other sections, (2) papers truly omitting limitations, and (3) extraction pipeline failures. The asterisks ($^*$) mark non-compliance with the mandatory reporting policy.}
    \label{tab:non_compliant_breakdown}
\end{table}

\begin{figure}[t]
    \centering
    \includegraphics[width=\linewidth]{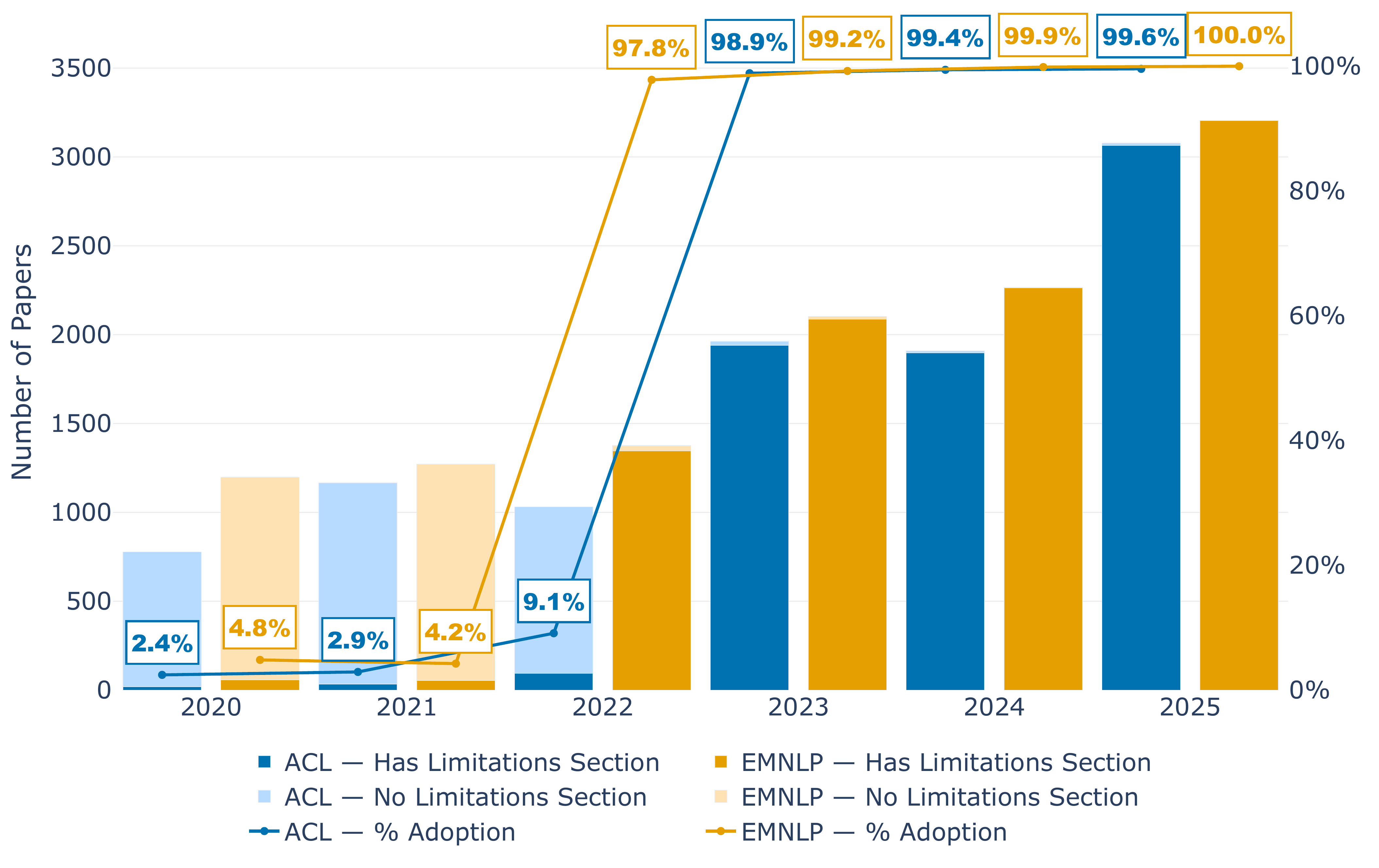} 
    \caption{The presence of explicit Limitations sections in ACL and EMNLP papers (2020--2025). Papers lacking a Limitations section from EMNLP 2022 onwards have been manually verified.}
    \label{fig:temporal_prevalence}
\end{figure}

\subsection{Prevalence of Limitations Sections}
\label{sec:data_stats}

To understand the impact of mandatory reporting policy, we analyzed the presence of explicit Limitations sections across ACL and EMNLP from 2020 to 2025.
Despite mandates being in place since EMNLP 2022, our extraction pipeline flagged 136 papers (0.85\%) accepted between EMNLP 2022 and EMNLP 2025 as lacking a required Limitations section.
To verify these, we manually inspected the 136 papers and reported the results in Table~\mbox{\ref{tab:non_compliant_breakdown}}. 
Among the 136 papers, we found that 41 papers (0.26\%) were flagged due to extraction pipeline failures, whereas 95 papers (0.60\%) genuinely missed a dedicated Limitations section.
Further analysis of these 95 papers reveals that 50 discussed their limitations within paragraphs embedded in other sections rather than in a required standalone section before the References. For the remaining 45 papers, we could not find any discussion paragraph, named Limitations, anywhere in the text. We believe these numbers are interesting information for future NLP conference organizers.

After the manual verification above, Figure \ref{fig:temporal_prevalence} plots the prevalence of the dedicated Limitations section over time. The trend shows that the policy mandates drove a dramatic shift in reporting practices. Prior to the mandates, voluntary inclusion was rare (less than 10\%). Afterward, compliance surged to 97.8\% at EMNLP 2022 and 98.9\% at ACL 2023, reaching perfect compliance (100\%) by EMNLP 2025.
This reflects how conference review processes gradually strengthened over time.

\subsection{RQ1: Temporal Analysis} \label{sec:rq1}

This section reports how the trends of reported limitations change over the six years we studied.

\begin{figure}[t]
    \centering
    \includegraphics[trim=0mm 0mm 0mm 10mm, clip, width=\columnwidth]{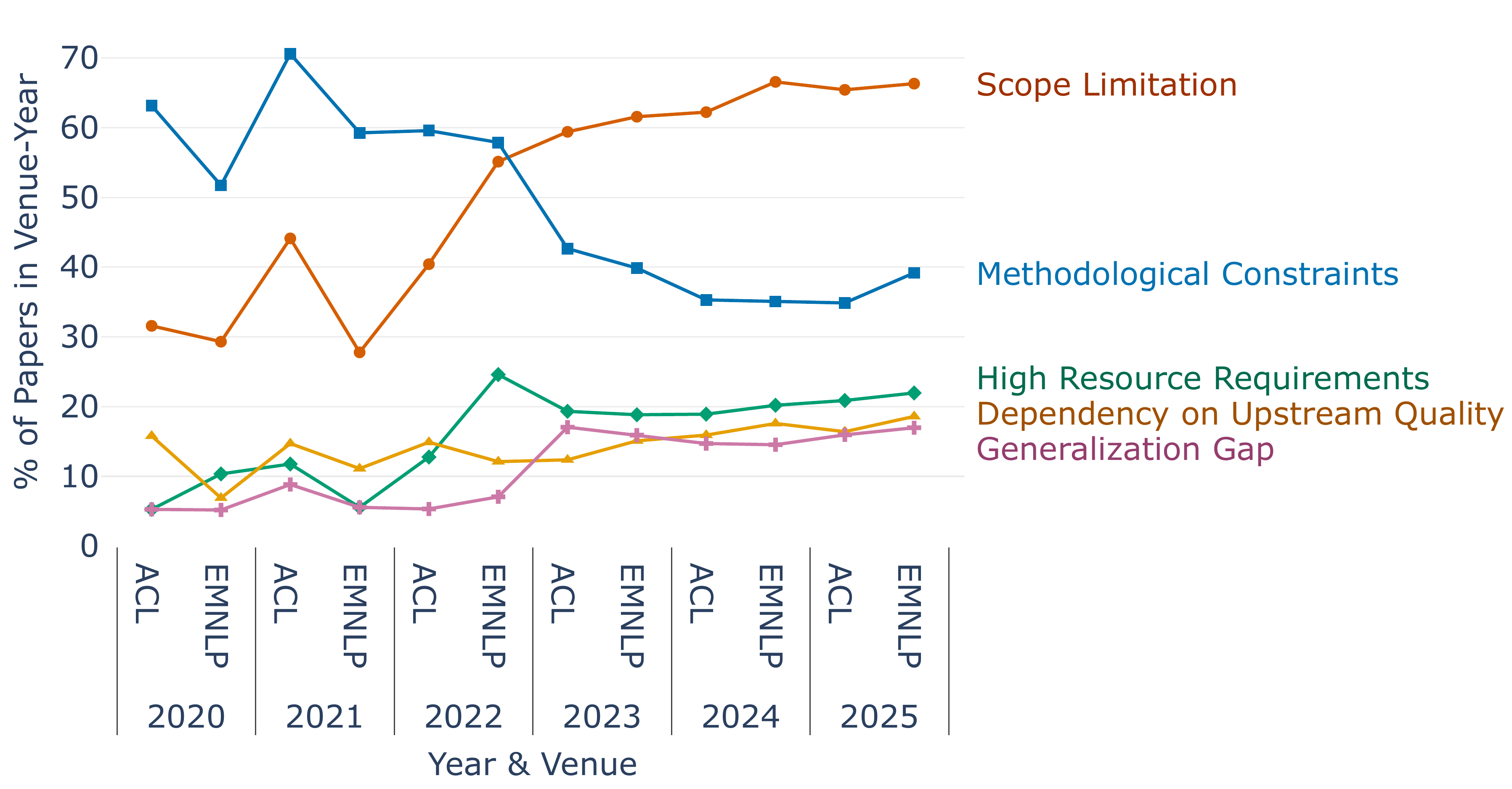}
    \caption{Trends of the top 5 limitation codes.}
    \label{fig:code_trends}
\end{figure}

\paragraph{Limitation Topics.}
Figure \ref{fig:code_trends} illustrates the evolving nature of self-reported limitations from 2020 to 2025. A striking divergence occurs between the two most prevalent codes: while \textit{Methodological Constraints} experienced a sharp decline after peaking in 2021, \textit{Scope Limitation} exhibits a massive upward trajectory, surging from a prevalence of $\sim$30\% in 2020 to over 65\% by 2025. 
Notably, the rankings of these two codes swapped almost precisely at EMNLP 2022, when the mandatory policy was first introduced. 
This intersection raises the compelling question of whether the policy mandate was the primary driver of this divergence or it resulted from other confounding factors.
A more controlled analysis is needed to answer this question.

\begin{figure}[t]
    \centering
    \includegraphics[trim=0mm 0mm 0mm 10mm, clip, width=\columnwidth]{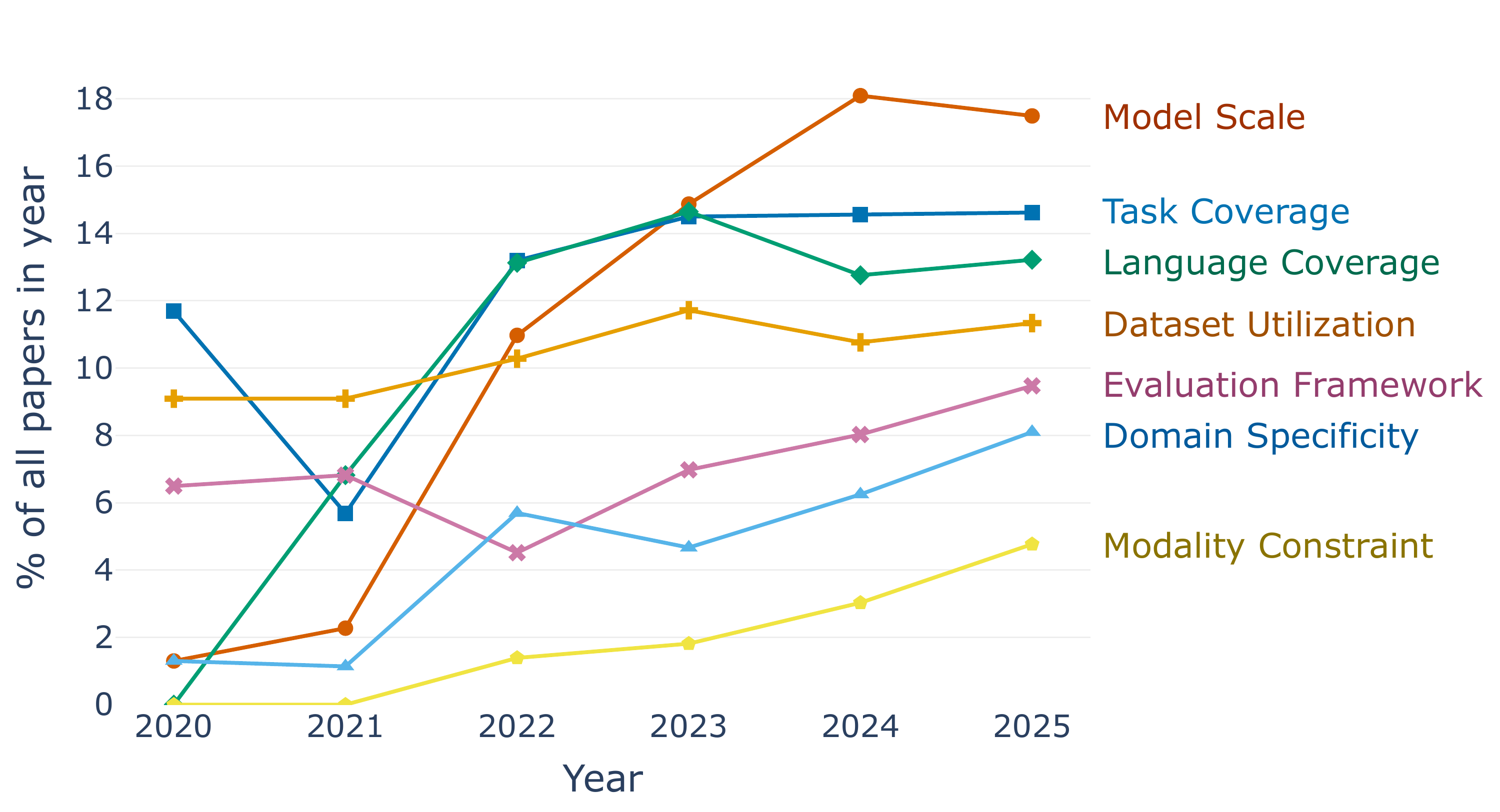}
    \caption{Trends of \textit{Scope Limitation} sub-codes.}
    \label{fig:scope_sub_code_trends}
\end{figure}

Driven by the increase in \textit{Scope Limitation}, we performed a second-level analysis on this code to investigate the underlying topics. Specifically, we employed LLM-driven clustering to construct a fine-grained sub-codebook, subsequently re-annotating the semantic units with \textit{Scope Limitation} to assign these sub-codes. As depicted in Figure \ref{fig:scope_sub_code_trends}, \textit{Model Scale} became the fastest-growing sub-code. This is typically characterized by researchers bounding their evaluated parameter sizes, frequently noting that they \textit{``limited our study to [Model X], and our findings may not generalize to larger models.''}. 
This reflects the recent paradigm shift brought by the advent of large language models.
In addition, we observed a sharp increase in the \textit{Language Coverage} sub-code during 2020-2022, likely stemming from growing awareness of the field's English-centric bias and recent calls to explicitly name studied languages \cite{bender2019benderrule}.

\paragraph{Word Counts and Code Diversity.} 
As reported in Table \ref{tab:word_count_and_codes}, the median length of Limitations sections dropped after the introduction of the mandatory policy (EMNLP 2022) and then stabilized at around 138--150 words. However, the average number of unique limitation codes per paper grew from 1.63 in ACL 2020 to 3.02 in EMNLP 2025. 
This reveals that authors are increasingly adopting a concise, multifaceted approach, i.e., packing more diverse limitation types into roughly the same space rather than elaborately describing a few types.

\begin{table}[t]
\small
\centering
\resizebox{\columnwidth}{!}{
\begin{tabular}{lcccccc}
\toprule
\multirow{2}{*}{\textbf{Year}} & \multicolumn{3}{c}{\textbf{ACL}} & \multicolumn{3}{c}{\textbf{EMNLP}}  \\
\cmidrule(lr){2-4} \cmidrule(lr){5-7}
& \textbf{Mean} & \textbf{Med.} & \textbf{\#Codes} & \textbf{Mean} & \textbf{Med.} & \textbf{\#Codes} \\
\midrule
2020 & 156.42 & 142.00 & 1.63 & 201.41 & 152.50 & 1.90 \\
2021 & 204.47 & 191.00 & 2.44 & 180.04 & 150.50 & 1.70 \\
2022 & 220.03 & 171.50 & 2.47 & 167.97 & 138.00 & 2.49 \\
2023 & 173.24 & 138.00 & 2.89 & 177.06 & 146.00 & 2.89 \\
2024 & 168.90 & 139.00 & 2.91 & 182.73 & 150.00 & 2.93 \\
2025 & 173.77 & 139.00 & 2.89 & 175.77 & 139.00 & 3.02 \\
\bottomrule
\end{tabular}%
}
\caption{Mean and median word counts alongside the average number of unique limitation codes (\#Codes) per Limitations section in each conference.}
\label{tab:word_count_and_codes}
\end{table}

\paragraph{Iterative Codebook Updates.}
Across the four iterations, the codebook was updated by 27 \textsc{Expand} actions and 25 \textsc{Add} actions. 
Out of the 25 new codes, 22 were added after the first iteration due to the limited coverage of our initial codebook. Besides that, a new code, namely \textit{Literature Coverage Gap}, was introduced in 2024 to capture authors discussing their inability to review all relevant literature, stating for instance: \textit{``However, it is possible that some other relevant works were overlooked...''}. Following this, two new codes were added in 2025: \textit{Dataset Task Mismatch} and \textit{Sparse Data Sensitivity}. 
At the end, the final version of our codebook has 40 limitation codes and 10 non-limitation codes. 
Details of the codebook evolution are provided in Appendix \ref{app:codebook_evolution}.

\subsection{RQ2: Correlations with Paper Attributes} \label{sec:rq2}
We cross-analyzed the limitation codes with three paper attributes: research area, paper format, and author affiliation. 

\paragraph{Research Area.} \label{par:research_area}

\begin{table}[t]
\centering
\footnotesize
\setlength{\tabcolsep}{4pt}
\begin{tabular}{@{}l r r@{}}
\toprule
\textbf{Research Area $\times$ Limitation Code} & \textbf{O/E} & \textbf{Adj. $p$} \\
\midrule
\multicolumn{3}{@{}l}{\textbf{Human-Centered NLP}} \\
\quad \textit{Societal and Ethical Risks} & 5.80 & $<$0.001 \\
\quad \textit{Subjectivity in Evaluation/Annotation} & 3.83 & $<$0.001 \\
\addlinespace[2pt]
\multicolumn{3}{@{}l}{\textbf{Efficient NLP}} \\
\quad \textit{Hyperparameter Sensitivity} & 3.53 & $<$0.001 \\
\addlinespace[2pt]
\multicolumn{3}{@{}l}{\textbf{Interpretability}} \\
\quad \textit{Lack of Interpretability} & 3.31 & $<$0.001 \\
\quad \textit{Theoretical Gap} & 2.77 & $<$0.001 \\
\addlinespace[2pt]
\multicolumn{3}{@{}l}{\textbf{Summarization}} \\
\quad \textit{Reliance on Automatic Metrics} & 3.29 & 0.002 \\
\addlinespace[2pt]
\multicolumn{3}{@{}l}{\textbf{Language Modeling}} \\
\quad \textit{Reproducibility Gap} & 2.91 & 0.036 \\
\addlinespace[2pt]
\multicolumn{3}{@{}l}{\textbf{Computational Social Science}} \\
\quad \textit{Temporal Degradation} & 2.90 & $<$0.001 \\
\addlinespace[2pt]
\multicolumn{3}{@{}l}{\textbf{Machine Translation}} \\
\quad \textit{Reliance on Automatic Metrics} & 2.82 & 0.006 \\
\addlinespace[2pt]
\multicolumn{3}{@{}l}{\textbf{Ethics, Bias, and Fairness}} \\
\quad \textit{Privacy and Security Risks} & 2.78 & 0.005 \\
\bottomrule
\end{tabular}
\caption{Top 10 significant (research area, limitation code) pairs with the highest observed-to-expected ratio (O/E). Adj. $p$ is the $p$-value of the residual analysis after Bonferroni correction.}
\label{tab:area_residuals}
\end{table}

To investigate whether reported limitations vary by research area, we analyzed 6,742 papers where research areas could be gathered from online conference programs.
Specifically, we conducted a chi-square test of homogeneity on a 29 $\times$ 40 contingency table (research areas $\times$ unique limitation codes). The test rejected the null hypothesis of equal distributions across areas ($\chi^{2}(1{,}092) = 2{,}872$, $p < 0.001$, Cram\'er's $V = 0.071$), indicating a medium effect size under Cohen's df-adjusted benchmarks. 
This means certain research areas reported specific limitations at rates significantly different from the field average.

To uncover such prominent areas and limitations, we ran a post-hoc analysis using standardized residuals. 
The results reveal that 32 out of 1,160 cells in the contingency table had significant residuals after the Bonferroni correction \cite{dunn1961multiple}.
Table~\mbox{\ref{tab:area_residuals}} lists the top 10 significant area-limitation pairs with the highest observed-to-expected ratio.
For example, Human-Centered NLP papers over-reported \textit{Societal and Ethical Risks}, whereas Efficient NLP papers over-reported \textit{Hyperparameter Sensitivity}.
These pairs reflect the distinct priorities or concerns of each subfield.
However, the 32 significant cells account for only 2.76\% of the contingency table. 
Thus, roughly 97\% of (research area, limitation code) pairs do not differ significantly from the field average, demonstrating a field-wide homogeneity in how limitations are reported. 
These findings motivate future work to investigate the underlying causes of this uniformity and to find out whether similar patterns exist in adjacent fields such as HCI or general machine learning.

\paragraph{Paper Format.}
We further investigated whether the publication format (Long, Short, or Findings)
influences limitation reporting. Here, we focused on 7,033 ACL 2021 to ACL 2025 papers, where the format
can be easily extracted from Anthology IDs (e.g., \texttt{2025.acl-long.1}).
Testing all limitation codes for independence from paper format, we found no differences that remain significant after Benjamini--Hochberg correction \cite{benjamini1995controlling}. 
However, a few codes show differences substantial enough to report as exploratory observations.
Short papers reported empirical limitation types more often than long and Findings
papers. For example, 12.2\% of short papers reported \textit{Empirical Underperformance},
whereas only 7.5\% of both long and Findings papers did so (Adjusted $p = 0.266$). 
Similarly, \textit{Hyperparameter Sensitivity} was found more frequently in short papers (4.5\%)
than in long (2.5\%) and Findings (2.9\%) papers (Adjusted $p = 0.503$). 
These are consistent with the inherent nature of short papers, which often present
targeted, smaller-scale experiments, negative results, or preliminary findings.
The full code distribution and per-code test results are provided in
Appendix~\ref{app:code_distribution_by_format}.

\paragraph{Author Affiliation.} 
We investigated whether research from large companies reports similar types of limitations as that from other institutions. Using the list of large companies from the Forbes Global 2000 (2025) dataset\footnote{\url{https://www.kaggle.com/datasets/ellimaaac/forbes-the-global-2000-companies-2025}}, we categorized papers into three groups: (1) papers with large company affiliations only, (2) papers with mixed affiliations, and (3) papers without large company affiliations. 
Because the influence of large companies on papers with mixed affiliations varies across individual papers, we focused on comparisons between papers with only large company affiliations and papers without large company affiliations.
Specifically, to compare the two groups, we conducted two-sided Fisher’s exact tests for each limitation code
and adjusted the resulting $p$-values using the Benjamini–Hochberg correction. 

\begin{figure}[t]
    \centering
    \includegraphics[width=\columnwidth]{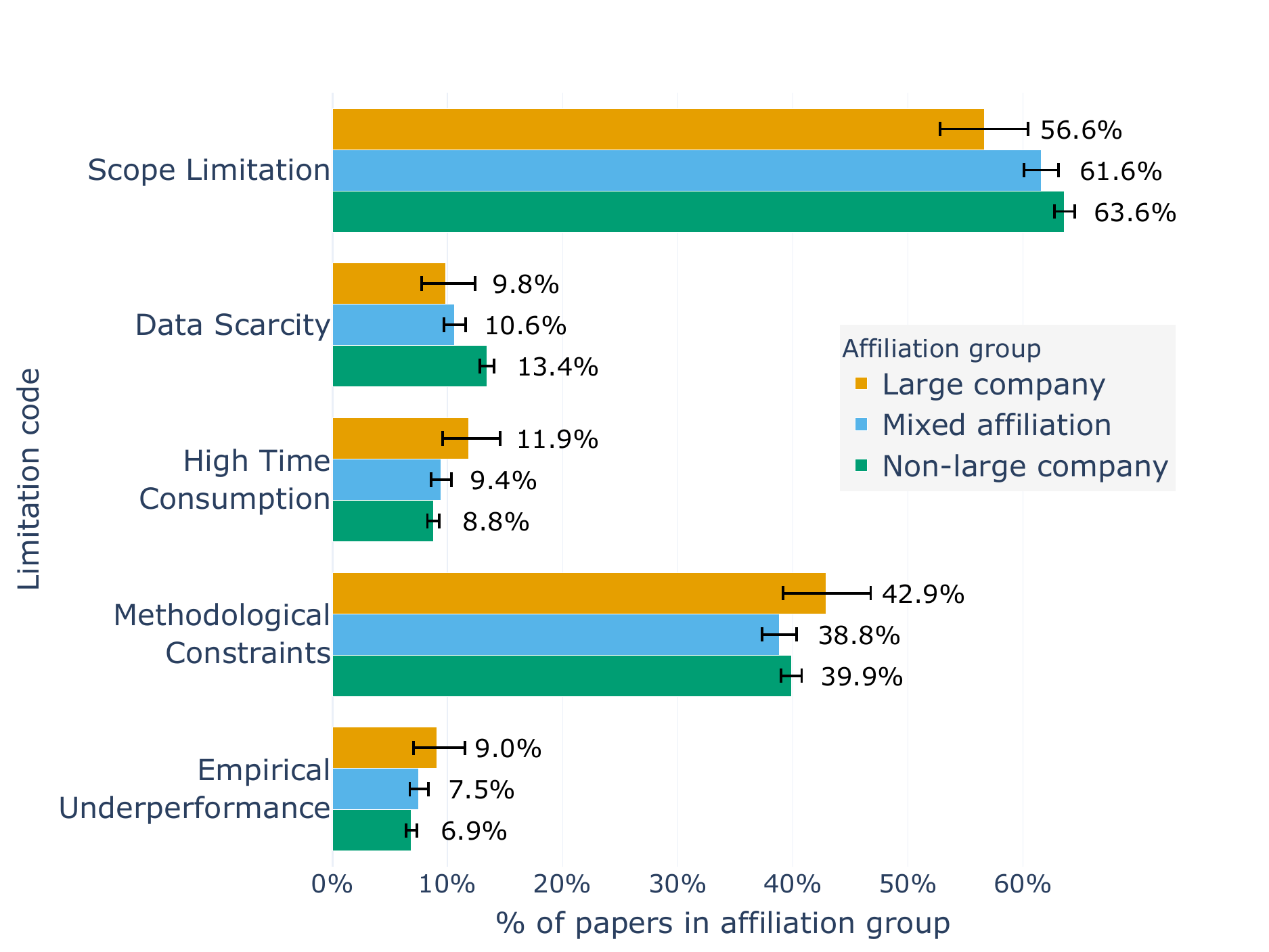}
    \caption{Top 5 limitation codes with the highest percentage difference between the Large company and Non-large company affiliation groups, ordered from highest to lowest difference. Bars show the percentage of papers within each affiliation group that mention the limitation code. Error bars represent the 95\% confidence intervals.
    }
    \label{fig:affiliation_fig}
\end{figure}

Figure \mbox{\ref{fig:affiliation_fig}} shows that papers without large company affiliations reported a significantly higher rate of \textit{Scope Limitation} compared to papers with only large company affiliations (63.6\% vs 56.6\%, Adjusted $p = 0.019$). 
They also exhibited a higher rate of \textit{Data Scarcity} (13.4\% vs. 9.8\%) although this difference did not reach statistical significance. 
Conversely, while papers with only large company affiliations exhibited higher raw percentages in codes such as 
\textit{High Time Consumption}, \textit{Methodological Constraints}, and \textit{Empirical Underperformance}, 
the differences were not statistically significant.
Full test results are detailed in Appendix\mbox{~\ref{app:affiliation_code_sig_test}}.

\subsection{RQ3: Discursive Patterns} \label{sec:rq3}

We systematically tracked the presence of ``Non-Limitation'' (NL) codes
to observe recurring discursive patterns in self-reported limitations.
These NL codes include, e.g., outlining \textit{Future Work}, giving a \textit{Contextual Justification}, highlighting \textit{Strong Reported Performance}, explaining \textit{Conducted Mitigation}, and providing \textit{Authorial Disclaimers}. 

\begin{table*}[t]
\centering
\footnotesize
\setlength{\tabcolsep}{6pt}
\begin{tabular}{l r r r r r r r r}
\toprule
& \multicolumn{4}{c}{\textbf{Lim$\to$NL}} & \multicolumn{4}{c}{\textbf{NL$\to$Lim}} \\ \cmidrule(lr){2-5} \cmidrule(l){6-9}
\textbf{NL Code} & \multicolumn{1}{r}{\textbf{Null}} & \multicolumn{1}{r}{\textbf{Observed}} & \multicolumn{1}{r}{\textbf{Diff}} & \multicolumn{1}{r}{\textbf{Adj. $p$}}  & \multicolumn{1}{r}{\textbf{Null}} & \multicolumn{1}{r}{\textbf{Observed}} & \multicolumn{1}{r}{\textbf{Diff}} & \multicolumn{1}{r}{\textbf{Adj. $p$}}\\
\midrule
Future Work & 32.8 & 40.2 & $+$7.4 & $<$0.001 & 32.8 & 27.9 & $-$4.9 & $<$0.001 \\
Contextual Justification & 17.0 & 19.7 & $+$2.7 & $<$0.001 & 17.0 & 17.3$^*$ & $+$0.2 & 0.280 \\
Conducted Mitigation & 6.3 & 8.2 & $+$2.0 & $<$0.001 & 6.3 & 5.8 & $-$0.4 & 0.002 \\
Method Details & 12.6 & 7.4 & $-$5.1 & $<$0.001 & 12.6 & 16.3 & $+$3.7 & $<$0.001 \\
Theoretical Projection & 6.3 & 7.3 & $+$1.0 & $<$0.001 & 6.3 & 4.8 & $-$1.5 & $<$0.001 \\
Strong Reported Performance & 11.8 & 4.5 & $-$7.3 & $<$0.001 & 11.8 & 17.5 & $+$5.7 & $<$0.001 \\
Method Strength & 4.7 & 3.8 & $-$1.0 & $<$0.001 & 4.7 & 4.9$^*$ & $+$0.1 & 0.234 \\
Recommendations & 2.8 & 3.5 & $+$0.7 & $<$0.001 & 2.8 & 1.9 & $-$0.9 & $<$0.001 \\
Anticipated Impact & 3.5 & 3.0 & $-$0.6 & $<$0.001 & 3.6 & 2.1 & $-$1.4 & $<$0.001 \\
Authorial Disclaimers & 2.2 & 2.4 & $+$0.2 & 0.003 & 2.2 & 1.6 & $-$0.6 & $<$0.001 \\
\bottomrule
\end{tabular}
\caption{Distribution of observed non-limitation codes as immediate successors
(Lim$\rightarrow$NL) and predecessors (NL$\rightarrow$Lim) of limitation codes.
Percentages are normalized over all transitions of each direction corpus-wide.
Null gives the expected baseline rate under 2{,}000 permutations that randomly shuffle the code sequence within each paper, preserving its marginal code frequencies. 
Diff equals the observed rate minus the null rate. 
All the observed rates, except those marked $^{\ast}$, significantly differ from the null according to the permutation tests at adjusted $p < 0.05$ after Benjamini--Hochberg correction.}
\label{tab:rhetorical_transitions}
\end{table*}

\paragraph{Implicit Limitation Reports.} \label{implicit-contraints}
While analyzing the LLM-annotated data, we surprisingly uncovered papers containing only non-limitation codes. For instance, we identified 90 such papers (1.4\%) in 2025. A manual review of these cases revealed sections in which the limitation appears only in the form of proposed future work. To illustrate, a paper might state, \textit{``Expanding beyond A and B remains an area for future exploration,''} without explicitly acknowledging that the current study was limited to evaluating only A and B. 
In such cases, the limitations are not formally declared, risking important limitations being overlooked.

\paragraph{Recurring Discursive Patterns.} 
\label{par:rq3_transition}
Next, we analyzed the relative placement of non-limitation (NL) and limitation (Lim) codes within these sections. 
While sequential proximity does not inherently imply a semantic relationship, certain non-random patterns are noteworthy because they align with potential rhetorical strategies (even if deliberate authorial intent cannot be confirmed).
Table~\ref{tab:rhetorical_transitions} compares the rates of NL codes immediately preceding or following a limitation code against baseline rates from a permutation null model (obtained by shuffling code sequences within each paper).
Among the NL codes immediately following a limitation code, 40.2\% are \textit{Future Work}, followed by \textit{Contextual Justification} (19.7\%), both of which are significantly higher than the null baselines. 
We label this textual pattern \textbf{the ``Soft Landing''}, which may help soften a limitation by suggesting future work or justifying why the limitation was unavoidable.

Additionally, 8.2\% of post-limitation NL codes are \textit{Conducted Mitigation}, detailing \textbf{proactive efforts} made by authors to address the stated limitation, regardless of whether these attempts were fully successful, partially successful, or unsuccessful. 
Certain limitation codes precede \textit{Conducted Mitigation} more frequently than others, indicating that authors often highlighted how they proactively addressed these issues:
\textit{Data Leakage/Contamination}, \textit{Reproducibility Gap}, and \textit{Subjectivity in Evaluation/Annotation}. 

Finally, we analyzed NL codes that precede limitation codes and found that 17.5\% of them are \textit{Strong Reported Performance} (SRP). We label this textual pattern the \textbf{preemptive buffer}, which may cushion a disclosure by preceding it with a statement of strong results. Notably, when the subsequent limitation is \textit{Empirical Underperformance}, this SRP buffer precedes it 23.3\% of the time. 

More details about this sequential analysis can be found in Appendices \ref{app:cm_vs_fw} and \ref{app:after_srp}.
Since these patterns were inferred solely from sequential orders, we encourage future studies to semantically analyze code relations for deeper insights.

\section{Conclusion}
This paper analyzes self-reported Limitations sections in NLP research, offering three core contributions: an iterative hybrid coding framework, a large-scale LLM-annotated dataset, and empirical findings addressing our three RQs. We hope these contributions help deepen understanding of the field
and spark community conversations on what constitutes a ``good'' limitation disclosure and how researchers can better use this mandatory section for transparent and responsible research.

\section*{Limitations}

\paragraph{Reliance on Explicit Self-Reporting.} 
Our study analyzed contents in the dedicated Limitations sections only. Consequently, our analysis did not include unwritten limitations as well as limitations dispersed elsewhere in the paper, such as within the Methodology, Discussion, or Conclusion sections. Additionally, because the analyzed data relied on self-reporting by the authors, our findings capture \textit{what is said} rather than \textit{what is true}. Future research investigating unreported or implicitly stated limitations would enrich this line of inquiry, allowing for a more holistic evaluation of research transparency.

\paragraph{Caveats on Causal Interpretation.} 
Several patterns we observed reflect salient correlations between limitation contents and paper attributes, such as publication year, research area, and author affiliations. However, these correlations do not inherently imply causation. For instance, we cannot definitively conclude that the mandatory section policy (first introduced at EMNLP 2022) caused the diverging trends of \textit{Scope Limitation} and \textit{Methodological Constraints} shown in Figure~\ref{fig:code_trends}. These shifts might instead be driven by confounding factors, such as the rapid adoption of large language models that occurred around the same period. Readers should therefore interpret these results cautiously.

\paragraph{Limited Context for Coding.}
A Limitations section is typically placed after the Conclusion, so authors may reference specific methodologies, issues, figures, or tables introduced in preceding sections. Sometimes, context from these earlier sections is essential for accurately assigning codes. 
Therefore, our pipeline, which takes only the Limitations section as input, may miss important contextual information.
While it performed relatively well, the remaining performance gap could likely be narrowed by incorporating preceding sections as additional input. Nonetheless, this context expansion should be implemented selectively to optimize token efficiency, such as by dynamically retrieving external sections only when required.


\appendix
\label{sec:appendix}
\onecolumn
\section{Per-Code Agreement Breakdown} \label{app:percode}
To examine how consistently the human annotators and $M_{\text{code}}$ assign each code,
we compute agreement separately for every code in the codebook. For each code, we treat
coding as a binary decision per paper and compute Krippendorff's $\alpha$ twice: once
over the independent pre-consensus labels of the human annotators assigned to each
paper, and once between $M_{\text{code}}$ and the adjudicated labels. The two quantities
use different references and are therefore not directly comparable. We report them side
by side to show where model difficulty and annotator difficulty coincide.
Table~\ref{tab:percode} lists all 25 codes, ordered by the Model vs. Adjudicated column. Support counts the papers in which $M_{\text{code}}$, the adjudicated labels, or
both assign the code, and so pertains to the rightmost column. The annotator column is
computed over its own support. 
At such low support, a small number
of disagreements is sufficient to shift $\alpha$ substantially in either direction. This
accounts for both the ceiling values observed at the top of the table
(\textit{Model Hallucination/Incoherence}, \textit{Potential for Misuse}) and the floor
value observed at the bottom (Among Annotators $\alpha$ is undefined for \textit{Reliance on Automatic Metrics} because fewer than two annotators coded any paper containing it.). Agreement values
for these codes should therefore be interpreted with caution. Two further codes,
\textit{Lack of Interpretability} and \textit{Data Leakage/Contamination}, occur in only
one or two papers, too few for $\alpha$ to be defined at all.

\begin{table}[ht]
\centering
\footnotesize
\setlength{\tabcolsep}{18pt}
\begin{tabular}{l r r r}
\toprule
\textbf{Code} & \textbf{Support} & \makecell[r]{\textbf{Among}\\\textbf{Annotators ($\alpha$)}} & \makecell[r]{\textbf{Model vs.}\\\textbf{Adjudicated ($\alpha$)}} \\
\midrule
Model Hallucination/Incoherence & 6 & 0.743 & 1.000 \\
High Time Consumption & 16 & 0.640 & 0.926 \\
High Resource Requirements & 24 & 0.817 & 0.922 \\
NL: Future Work & 112 & 0.822 & 0.868 \\
Subjectivity in Evaluation/Annotation & 11 & 0.247 & 0.832 \\
Potential for Misuse & 3 & 0.797 & 0.797 \\
NL: Strong Reported Performance & 43 & 0.512 & 0.765 \\
Scope Limitation & 94 & 0.637 & 0.717 \\
NL: Conducted Mitigation & 16 & 0.423 & 0.696 \\
Lack of Evaluation Metrics or Benchmarks & 6 & 0.885 & 0.658 \\
Data Scarcity & 16 & 0.434 & 0.639 \\
Dataset Bias/Imbalance & 20 & 0.479 & 0.631 \\
NL: Contextual Justification & 60 & 0.497 & 0.616 \\
Reliance on External Tools or Resources & 9 & 0.427 & 0.599 \\
High Financial Cost & 7 & 0.321 & 0.588 \\
NL: Method Details & 47 & 0.415 & 0.531 \\
NL: Method Strength & 29 & 0.455 & 0.521 \\
NL: Anticipated Impact & 11 & 0.656 & 0.510 \\
Dependency on Upstream Quality & 19 & 0.763 & 0.496 \\
Methodological Constraints & 81 & 0.341 & 0.494 \\
Low Data Quality & 10 & 0.407 & 0.439 \\
Performance Trade-off & 5 & 0.230 & 0.322 \\
Reliance on Automatic Metrics & 3 & --- & $-0.007$ \\
\midrule
Lack of Interpretability & 2 & --- & --- \\
Data Leakage/Contamination & 1 & --- & --- \\
\bottomrule
\end{tabular}
\caption{Per-code Krippendorff's $\alpha$ on the 150 annotated papers, ordered by agreement with the adjudicated labels. ``NL:'' marks non-limitation codes; ``---'' marks codes with too few instances for $\alpha$ to be defined.}
\label{tab:percode}
\end{table}
\section{Second-Level Analysis of Scope Limitation}
\label{appendix:second_level_scope}
Driven by the significant trend increase in the prevalence of \textit{Scope Limitation}, we investigated the underlying sub-topics driving its mentions. Specifically, we developed a sub-codebook to re-annotate the semantic units previously assigned to this \textit{Scope Limitation} code as explained in Section~\ref{sec:rq1}. Definitions of the sub-codes are detailed below:

\begin{itemize}
    \item \textbf{Model Scale:} Grouping disclosures that limit the experimental evaluation to specific model sizes, parameter counts, or distinct model families, frequently excluding larger or proprietary closed-source LLMs due to strict computational constraints; exemplified by: \textit{Our experimental scope was constrained to backbone models under 8 billion parameters due to computational limitations.''} 
    \item \textbf{Task Coverage:} Restricts the operational or evaluation scope to simple, single-turn, single-hop, or short-form tasks, explicitly omitting more complex, multi-step, or long-form task structures; exemplified by: \textit{Our investigation is based on short-form (sentence-length) QA datasets, which may not fully capture the complexity of real-world scenarios.''}
    \item \textbf{Language Coverage:} Limits the boundaries of the study to a specific language (predominantly English) or a highly restricted subset of languages, thereby formally acknowledging the lack of broader multilingual generalization; exemplified by: \textit{Additionally, our models have not been examined on languages beyond English.''} 
    \item \textbf{Dataset Utilization:} Confines the empirical evaluation to a single dataset, a narrow subset of data, or specific localized benchmark collections rather than diverse, large-scale benchmarks; exemplified by: \textit{A clear limitation of this work is that it exclusively focuses on a single dataset.''}
    \item \textbf{Evaluation Framework:} Restricts the methodology to specific testing paradigms, such as zero-shot evaluation or prompting-only setups, while explicitly excluding fine-tuning, model adaptation, or extensive hyperparameter optimization; exemplified by: \textit{Evaluation is performed in a prompting-only setup without model adaptation or tuning.''} 
    \item \textbf{Domain Specificity:} Confines the research questions, datasets, or evaluation environments to highly specialized fields, vertical topics, or distinct vertical domains; exemplified by: \textit{Our research exclusively focuses on the task of factuality alignment in clinical summarization.''}
    \item \textbf{Modality Constraint:} Characterized by a text-only design paradigm that explicitly excludes alternative modalities—such as audio, vision, or speech—to focus entirely on textual data processing; exemplified by: \textit{``The benchmark is limited to text and does not include multimodal inputs such as vision or speech.''}
\end{itemize}

The yearly distributions of these sub-codes of \textit{Scope Limitation} are shown in Table~\ref{tab:scope_subcode_distribution}.

\begin{table*}[h]
\centering
\small
\resizebox{\textwidth}{!}{%
\begin{tabular}{lrrrrrr}
\toprule
\textbf{Sub-code} & \textbf{2020} & \textbf{2021} & \textbf{2022} & \textbf{2023} & \textbf{2024} & \textbf{2025} \\
\midrule
Model Scale          & 1.3\% (1) & 2.3\% (2) & 11.0\% (158) & 14.9\% (599) & 18.1\% (753) & 17.5\% (1,097) \\
Task Coverage        & 11.7\% (9) & 5.7\% (5) & 13.2\% (190) & 14.5\% (584) & 14.6\% (606) & 14.6\% (917)  \\
Language Coverage    & 0.0\% (0) & 6.8\% (6) & 13.1\% (189) & 14.6\% (590) & 12.8\% (531) & 13.2\% (829)  \\
Dataset Utilization  & 9.1\% (7) & 9.1\% (8) & 10.3\% (148) & 11.7\% (472) & 10.8\% (448) & 11.3\% (711)  \\
Evaluation Framework & 6.5\% (5) & 6.8\% (6) & 4.5\% (65)   & 7.0\% (281)  & 8.0\% (334)  & 9.5\% (594)  \\
Domain Specificity   & 1.3\% (1) & 1.1\% (1) & 5.7\% (82)   & 4.7\% (188)  & 6.2\% (260)  & 8.1\% (508)  \\
Modality Constraint  & 0.0\% (0) & 0.0\% (0) & 1.4\% (20)   & 1.8\% (73)   & 3.0\% (126)  & 4.8\% (299)  \\
\bottomrule
\end{tabular}%
}
\caption{Yearly distributions of sub-codes within the \textit{Scope Limitation} code.}
\label{tab:scope_subcode_distribution}
\end{table*}
\newpage

\section{Code Distribution by Research Area}
Following the analysis of limitation codes and research areas in Appendix~\ref{app:research_area_clustering}, Figure~\ref{fig:code_distribution_by_area} shows the prevalence of different limitation codes across the research areas in NLP.

\begin{figure*}[h]
    \centering
    \includegraphics[width=\linewidth]{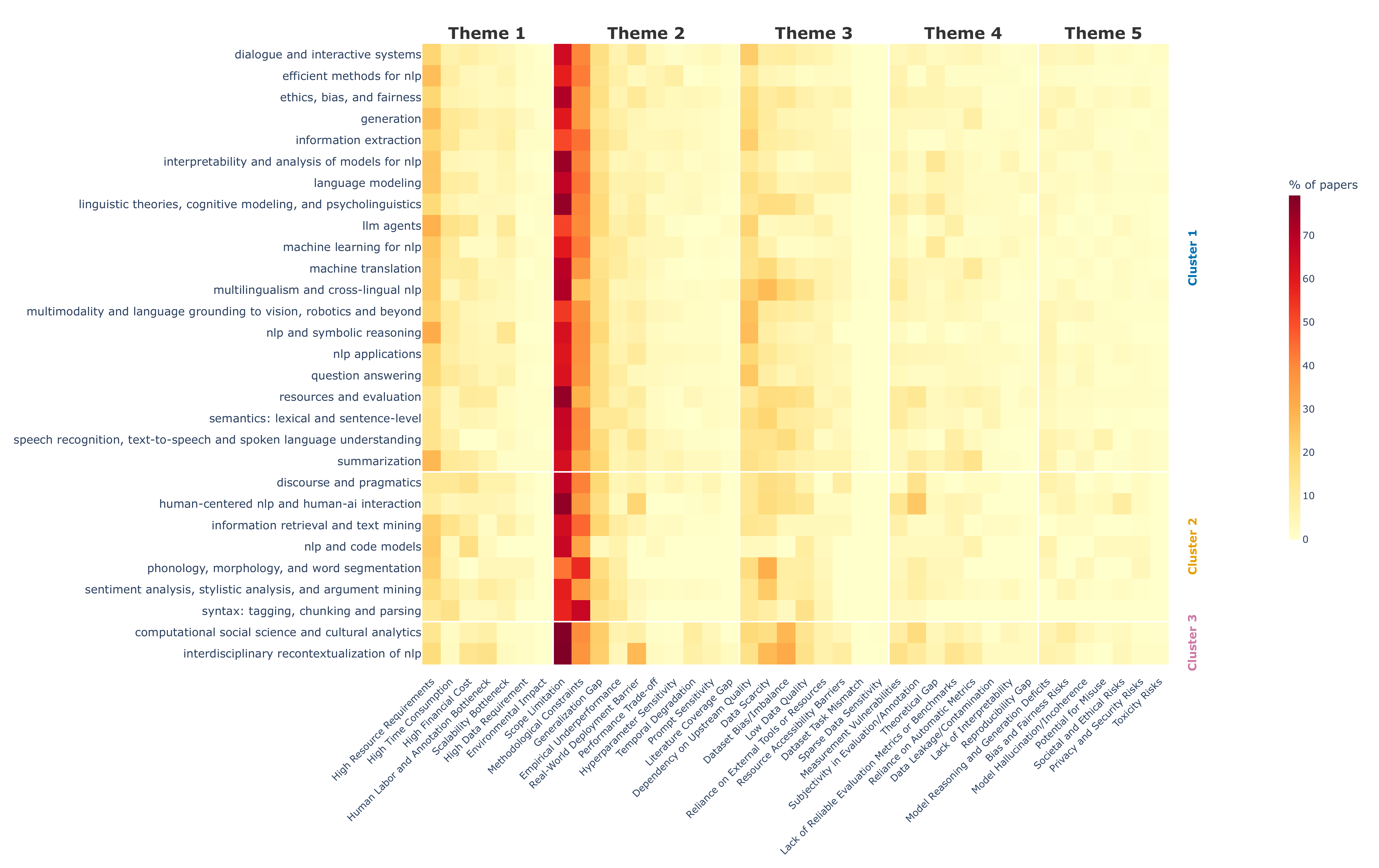}
    \caption{
    Distributions of limitation codes by NLP research areas, computed from 6,742 papers where research areas could be gathered from online conference programs. The research areas are grouped by area clusters in Appendix~\ref{app:research_area_clustering}, while the limitation codes are sorted by themes and corpus-wide prevalence discussed in Appendix~\ref{app:codebook}. 
    }
    \label{fig:code_distribution_by_area}
\end{figure*}
\clearpage
\section{Exploratory Clustering of Research Areas}
\label{app:research_area_clustering}
\begin{figure*}[ht]
    \centering
    \includegraphics[width=\linewidth]{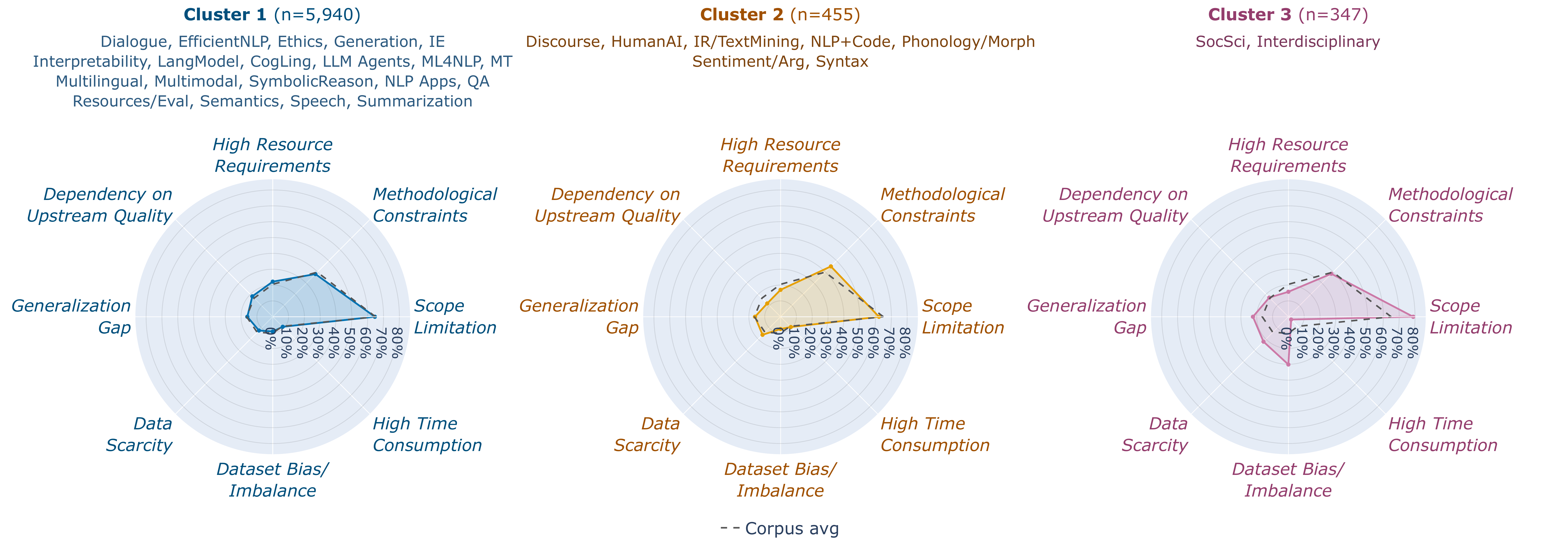}
    \caption{Limitation profiles of the three research area clusters with respect to the top-8 limitation codes, where the percentage axes indicate the proportion of papers within each cluster that report each specific code. The dashed grey lines represent the corpus-wide average profile.}
    \label{fig:track_clusters}
\end{figure*}

Before settling on the cell-level analysis reported in Section~\ref{sec:rq2}, we
attempted to group research areas into clusters by their limitation profiles. Each area
$q$ was represented as a $d$-dimensional vector $\mathbf{v}$, where $d$ is the number of
unique limitation codes in our final codebook and each element $v_c$ is the proportion of
papers in area $q$ assigned code $c$. Each dimension was normalized to $[0,1]$ using
Min-Max scaling, and we applied Agglomerative Hierarchical Clustering
\citep{müllner2011modernhierarchicalagglomerativeclustering}, selecting the number of
clusters using the Silhouette Score together with intra-/inter-cluster average
distances. The resulting partition is weak: the clustering in
Figure~\ref{fig:track_clusters} places roughly 88\% of papers in a single cluster that
nearly reproduces the corpus-wide profile. This is consistent with the near-uniform reporting found at the cell level in Section~\ref{sec:rq2}. Read together, the two analyses point to the same picture, that research areas report limitations in broadly similar ways.

\clearpage
\section{Detailed Code Distribution by Paper Format}
\label{app:code_distribution_by_format}
We leveraged the paper ID naming conventions in the ACL Anthology to determine the publication format (i.e., Long, Short, or Findings) for each paper. Specifically, 
we isolated and analyzed a subset of 7,033 ACL 2021 to ACL 2025 papers to investigate the macro-level distribution of limitation codes across these distinct publication formats. 
For each limitation code, we tested whether code presence is independent of publication format using a chi-square test on the corresponding $3 \times 2$ contingency table, reporting Cram\'er's V as the effect size and applying
Benjamini--Hochberg correction across all tests. The normalized distribution of these
codes and the per-code test results are presented in Table~\ref{tab:code_distribution_by_type}.
\begin{table}[ht]
\centering
\footnotesize
\setlength{\tabcolsep}{4pt}
\begin{tabular}{>{\raggedright\arraybackslash}p{0.30\textwidth} c c c r c r r}
\toprule
\textbf{Limitation Code} & \textbf{Long} & \textbf{Short} & \textbf{Findings} &
\textbf{$\chi^{2}(2)$} & \textbf{$p$} & \textbf{Cram\'er's V} & \textbf{Adj. $p$} \\
 & ($n=3{,}426$) & ($n=337$) & ($n=3{,}270$) & & & & \\
\midrule
Scope Limitation & 63.2\% & 64.7\% & 61.4\% & 3.02 & 0.221 & 0.021 & 0.637 \\
Methodological Constraints & 38.6\% & 37.1\% & 36.6\% & 2.92 & 0.232 & 0.020 & 0.637 \\
High Resource Requirements & 20.6\% & 18.7\% & 19.0\% & 2.81 & 0.245 & 0.020 & 0.637 \\
Generalization Gap & 15.6\% & 16.0\% & 15.8\% & 0.09 & 0.958 & 0.003 & 0.983 \\
Dependency on Upstream Quality & 14.5\% & 12.2\% & 16.1\% & 5.67 & 0.059 & 0.028 & 0.503 \\
Data Scarcity & 12.1\% & 14.5\% & 12.0\% & 1.84 & 0.399 & 0.016 & 0.874 \\
Dataset Bias/Imbalance & 10.9\% & 10.7\% & 9.4\% & 4.54 & 0.103 & 0.025 & 0.503 \\
High Time Consumption & 9.4\% & 7.7\% & 8.8\% & 1.45 & 0.485 & 0.014 & 0.874 \\
Low Data Quality & 7.8\% & 8.0\% & 8.1\% & 0.15 & 0.930 & 0.005 & 0.980 \\
\textbf{Empirical Underperformance} & \textbf{7.5\%} & \textbf{12.2\%} & \textbf{7.5\%} & \textbf{9.98} & \textbf{0.007} & \textbf{0.038} & \textbf{0.266} \\
Real-World Deployment Barrier & 7.3\% & 7.4\% & 6.7\% & 0.91 & 0.635 & 0.011 & 0.874 \\
Measurement Vulnerabilities & 6.8\% & 5.9\% & 6.4\% & 0.75 & 0.687 & 0.010 & 0.874 \\
High Financial Cost & 6.4\% & 5.0\% & 6.5\% & 1.07 & 0.587 & 0.012 & 0.874 \\
Human Labor and Annotation Bottleneck & 5.7\% & 6.8\% & 5.9\% & 0.72 & 0.699 & 0.010 & 0.874 \\
Subjectivity in Evaluation/Annotation & 5.6\% & 6.5\% & 5.1\% & 1.72 & 0.424 & 0.016 & 0.874 \\
Reliance on External Tools or Resources & 5.1\% & 4.2\% & 5.1\% & 0.61 & 0.738 & 0.009 & 0.874 \\
Model Reasoning and Generation Deficits & 5.3\% & 4.7\% & 4.7\% & 1.21 & 0.547 & 0.013 & 0.874 \\
Scalability Bottleneck & 4.5\% & 3.0\% & 5.1\% & 3.62 & 0.164 & 0.023 & 0.637 \\
Theoretical Gap & 5.1\% & 5.9\% & 4.1\% & 4.64 & 0.098 & 0.026 & 0.503 \\
Lack of Reliable Evaluation Metrics or Benchmarks & 4.6\% & 3.9\% & 4.1\% & 1.26 & 0.532 & 0.013 & 0.874 \\
Resource Accessibility Barriers & 4.3\% & 3.6\% & 3.5\% & 3.01 & 0.222 & 0.021 & 0.637 \\
Performance Trade-off & 3.6\% & 2.1\% & 3.1\% & 3.18 & 0.204 & 0.021 & 0.637 \\
Reliance on Automatic Metrics & 3.6\% & 1.5\% & 3.0\% & 5.36 & 0.069 & 0.028 & 0.503 \\
Bias and Fairness Risks & 3.2\% & 1.8\% & 2.8\% & 2.42 & 0.299 & 0.019 & 0.728 \\
\textbf{Hyperparameter Sensitivity} & \textbf{2.5\%} & \textbf{4.5\%} & \textbf{2.9\%} & \textbf{4.82} & \textbf{0.090} & \textbf{0.026} & \textbf{0.503} \\
Temporal Degradation & 2.5\% & 2.4\% & 2.4\% & 0.01 & 0.996 & 0.001 & 0.996 \\
Model Hallucination/Incoherence & 2.2\% & 2.7\% & 2.4\% & 0.54 & 0.762 & 0.009 & 0.874 \\
Prompt Sensitivity & 1.8\% & 2.1\% & 2.1\% & 1.15 & 0.562 & 0.013 & 0.874 \\
Data Leakage/Contamination & 1.8\% & 2.1\% & 1.9\% & 0.31 & 0.856 & 0.007 & 0.927 \\
High Data Requirement & 1.6\% & 1.5\% & 1.9\% & 0.65 & 0.721 & 0.010 & 0.874 \\
Toxicity Risks & 1.2\% & 0.0\% & 0.8\% & 7.18 & 0.028 & 0.032 & 0.503 \\
Privacy and Security Risks & 1.2\% & 0.9\% & 1.4\% & 0.83 & 0.660 & 0.011 & 0.874 \\
Reproducibility Gap & 1.2\% & 0.9\% & 1.3\% & 0.78 & 0.676 & 0.011 & 0.874 \\
Lack of Interpretability & 1.6\% & 1.8\% & 1.4\% & 0.60 & 0.741 & 0.009 & 0.874 \\
Potential for Misuse & 1.6\% & 1.5\% & 1.4\% & 0.46 & 0.795 & 0.008 & 0.886 \\
Societal and Ethical Risks & 1.9\% & 1.5\% & 1.4\% & 2.83 & 0.243 & 0.020 & 0.637 \\
Literature Coverage Gap & 0.4\% & 0.0\% & 0.8\% & 5.82 & 0.055 & 0.029 & 0.503 \\
Environmental Impact & 0.7\% & 0.6\% & 0.6\% & 0.59 & 0.745 & 0.009 & 0.874 \\
Sparse Data Sensitivity & 0.0\% & 0.0\% & 0.0\% & 1.05 & 0.591 & 0.012 & 0.874 \\
\bottomrule
\end{tabular}
\caption{Normalized distribution of the limitation codes across Long, Short, and
Findings papers, with a chi-square test of independence between code presence and
publication format for each code. 
\textit{Dataset Task Mismatch} does not occur in this subset and is therefore excluded. 
Cram\'er's V is the effect size, and we used the
Benjamini--Hochberg correction to adjust $p$-value across all the tests. Codes in \textbf{bold} are those
discussed in the \textit{Paper Format} paragraph of Section~\ref{sec:rq2}.}
\label{tab:code_distribution_by_type}
\end{table}

\newpage
\section{Affiliation Significance Test}
\label{app:affiliation_code_sig_test}

We compare the prevalence of each limitation code between papers classified as Large
company and Non-large company affiliation groups. Percentages are reported with 95\%
confidence intervals. The percentage-point difference ($\Delta$ pp) is calculated as the
Large-company percentage minus the Non-large-company percentage. Odds ratios (OR) and
two-sided Fisher's exact-test $p$-values were calculated for each code, and the reported
$p$-values were adjusted using the Benjamini-Hochberg correction. Mixed-affiliation papers
were excluded from this comparison.

\begin{table*}[ht]
\centering
\small
\setlength{\tabcolsep}{3pt}
\renewcommand{\arraystretch}{1.15}
\begin{tabular}{@{}>{\raggedright\arraybackslash}p{0.34\textwidth}p{0.18\textwidth}p{0.18\textwidth}rrr@{}}
\toprule
\textbf{Code} & \textbf{Large: \% [CI]} & \textbf{Non-large: \% [CI]} & \textbf{$\Delta$ pp} & \textbf{OR} & \textbf{Adj. $p$} \\
\midrule
Scope Limitation & 56.6\% [52.8, 60.4] & 63.6\% [62.7, 64.5] & $-$6.963 & 0.748 & 0.019 \\
High Time Consumption & 11.9\% [9.6, 14.6] & 8.8\% [8.2, 9.3] & 3.103 & 1.402 & 0.135 \\
Data Scarcity & 9.8\% [7.8, 12.4] & 13.4\% [12.8, 14.1] & $-$3.593 & 0.703 & 0.135 \\
Reliance on Automatic Metrics & 5.3\% [3.8, 7.3] & 3.4\% [3.1, 3.7] & 1.914 & 1.596 & 0.145 \\
Empirical Underperformance & 9.0\% [7.1, 11.5] & 6.9\% [6.4, 7.3] & 2.188 & 1.351 & 0.293 \\
Subjectivity in Evaluation/Annotation & 3.9\% [2.7, 5.7] & 5.8\% [5.4, 6.3] & $-$1.939 & 0.655 & 0.293 \\
Low Data Quality & 6.2\% [4.6, 8.4] & 8.4\% [7.9, 8.9] & $-$2.117 & 0.730 & 0.369 \\
Environmental Impact & 1.1\% [0.5, 2.2] & 0.6\% [0.4, 0.7] & 0.520 & 1.918 & 0.540 \\
Methodological Constraints & 42.9\% [39.1, 46.8] & 39.9\% [39.0, 40.8] & 3.043 & 1.134 & 0.571 \\
Hyperparameter Sensitivity & 3.6\% [2.4, 5.3] & 2.7\% [2.4, 3.0] & 0.902 & 1.348 & 0.571 \\
Model Reasoning and Generation Deficits & 3.3\% [2.2, 5.0] & 4.5\% [4.1, 4.9] & $-$1.206 & 0.722 & 0.571 \\
Dataset Bias/Imbalance & 9.2\% [7.2, 11.7] & 11.0\% [10.4, 11.6] & $-$1.804 & 0.820 & 0.571 \\
Dataset Task Mismatch & 0.2\% [0.0, 0.9] & 0.0\% [0.0, 0.1] & 0.121 & 4.434 & 0.574 \\
Temporal Degradation & 1.7\% [1.0, 3.0] & 2.5\% [2.3, 2.8] & $-$0.820 & 0.671 & 0.574 \\
Theoretical Gap & 3.6\% [2.4, 5.3] & 4.6\% [4.3, 5.0] & $-$1.053 & 0.765 & 0.574 \\
Resource Accessibility Barriers & 2.8\% [1.8, 4.4] & 3.9\% [3.6, 4.3] & $-$1.093 & 0.712 & 0.574 \\
Measurement Vulnerabilities & 5.3\% [3.8, 7.3] & 6.6\% [6.2, 7.1] & $-$1.327 & 0.789 & 0.574 \\
Lack of Reliable Evaluation Metrics or Benchmarks & 3.4\% [2.3, 5.1] & 4.4\% [4.1, 4.8] & $-$1.006 & 0.765 & 0.610 \\
High Financial Cost & 7.2\% [5.4, 9.4] & 6.2\% [5.8, 6.7] & 0.976 & 1.170 & 0.622 \\
Performance Trade-off & 4.1\% [2.8, 5.9] & 3.3\% [3.0, 3.6] & 0.754 & 1.238 & 0.622 \\
Bias and Fairness Risks & 3.4\% [2.3, 5.1] & 2.8\% [2.5, 3.1] & 0.640 & 1.238 & 0.622 \\
Data Leakage/Contamination & 2.3\% [1.4, 3.8] & 1.8\% [1.6, 2.1] & 0.508 & 1.284 & 0.634 \\
Reliance on External Tools or Resources & 4.2\% [2.9, 6.1] & 5.1\% [4.8, 5.6] & $-$0.931 & 0.811 & 0.634 \\
High Data Requirement & 1.7\% [1.0, 3.0] & 1.4\% [1.2, 1.7] & 0.298 & 1.214 & 0.774 \\
Real-World Deployment Barrier & 6.6\% [4.9, 8.7] & 7.4\% [6.9, 7.9] & $-$0.863 & 0.876 & 0.774 \\
Generalization Gap & 14.7\% [12.1, 17.6] & 15.7\% [15.0, 16.4] & $-$1.038 & 0.923 & 0.774 \\
Model Hallucination/Incoherence & 2.0\% [1.2, 3.4] & 2.5\% [2.2, 2.8] & $-$0.447 & 0.816 & 0.887 \\
Lack of Interpretability & 1.6\% [0.8, 2.8] & 1.9\% [1.7, 2.2] & $-$0.342 & 0.817 & 0.934 \\
Privacy and Security Risks & 1.2\% [0.6, 2.4] & 1.1\% [0.9, 1.3] & 0.165 & 1.154 & 0.957 \\
Reproducibility Gap & 1.2\% [0.6, 2.4] & 1.2\% [1.0, 1.4] & 0.086 & 1.075 & 0.958 \\
Toxicity Risks & 0.8\% [0.3, 1.8] & 0.7\% [0.6, 0.9] & 0.032 & 1.042 & 0.958 \\
Literature Coverage Gap & 0.5\% [0.2, 1.4] & 0.6\% [0.5, 0.8] & $-$0.157 & 0.747 & 0.958 \\
Prompt Sensitivity & 1.9\% [1.1, 3.2] & 2.1\% [1.8, 2.3] & $-$0.180 & 0.911 & 0.958 \\
Potential for Misuse & 1.4\% [0.7, 2.6] & 1.6\% [1.4, 1.8] & $-$0.181 & 0.884 & 0.958 \\
Societal and Ethical Risks & 1.2\% [0.6, 2.4] & 1.5\% [1.3, 1.7] & $-$0.258 & 0.827 & 0.958 \\
High Resource Requirements & 19.3\% [16.5, 22.6] & 19.7\% [19.0, 20.4] & $-$0.347 & 0.978 & 0.958 \\
Human Labor and Annotation Bottleneck & 5.1\% [3.7, 7.1] & 5.5\% [5.1, 5.9] & $-$0.356 & 0.932 & 0.958 \\
Scalability Bottleneck & 4.4\% [3.0, 6.2] & 4.6\% [4.2, 5.0] & $-$0.185 & 0.958 & 0.971 \\
Dependency on Upstream Quality & 16.1\% [13.4, 19.1] & 16.1\% [15.4, 16.8] & 0.005 & 1.000 & 1.000 \\
Sparse Data Sensitivity & 0.0\% [0.0, 0.6] & 0.0\% [0.0, 0.1] & $-$0.018 & 0.000 & 1.000 \\
\bottomrule
\end{tabular}
\caption{Proportions of reported limitation codes by Large company affiliation group
and Non-large company affiliation group.}
\label{tab:affiliation-limitation-tests}
\end{table*}

\newpage
\section{Transition Frequencies (Conducted Mitigation vs. Future Work)}
\label{app:cm_vs_fw}
Driven by our statistical analysis of discursive patterns, we observed that certain non-limitation (NL) codes follow specific limitation disclosures with significantly higher regularity than others. In particular, we focused on \textit{Conducted Mitigation}, which records steps already taken toward a stated limitation, encompassing successful, partially successful, and unsuccessful interventions. For each limitation code, we compared the share of its outgoing transitions that land on \textit{Conducted Mitigation} against the same permutation null used in
Section~\ref{sec:rq3}. Table~\ref{tab:transition_probabilities} reports the seven
codes with the highest observed share, all of which exceed their null and occasionally
surpass the prevalence of \textit{Future Work}, the most common NL code corpus-wide.

\begin{table*}[h]
\centering
\small
\setlength{\tabcolsep}{4pt}
\begin{tabular}{l r rrr rrr}
\toprule
\multirow{2}{*}{\textbf{Limitation Code}} & \multirow{2}{*}{\textbf{Total}} &
\multicolumn{3}{c}{\textbf{NL: Future Work}} &
\multicolumn{3}{c}{\textbf{NL: Conducted Mitigation}} \\
\cmidrule(lr){3-5} \cmidrule(lr){6-8}
& & \textbf{\%} & \textbf{Null \%} & \textbf{Adj.\ $p$}
& \textbf{\%} & \textbf{Null \%} & \textbf{Adj.\ $p$} \\
\midrule
Data Leakage/Contamination & 141 & 10.6 & 20.6 & 0.007 & \textbf{24.1} & 12.1 & \textbf{0.002} \\
Reproducibility Gap & 68 & 11.8 & 22.0 & 0.072 & \textbf{23.5} & 12.7 & \textbf{0.005} \\
Subjectivity in Evaluation/Annotation & 364 & 23.1 & 25.9 & 0.402 & \textbf{23.4} & 11.0 & \textbf{0.002} \\
Reliance on Automatic Metrics & 259 & 32.8 & 30.2 & 0.484 & 20.8 & 9.9 & 0.027 \\
Prompt Sensitivity & 130 & 27.7 & 29.2 & 0.845 & 20.0 & 9.3 & 0.025 \\
Privacy and Security Risks & 80 & 27.5 & 27.7 & 0.993 & 20.0 & 7.9 & 0.039 \\
Toxicity Risks & 57 & 7.0 & 26.0 & 0.007 & \textbf{19.3} & 9.0 & \textbf{0.002} \\
\bottomrule
\end{tabular}
\caption{Transition frequencies from specific limitation codes to two non-limitation
codes, \textit{Future Work} and \textit{Conducted Mitigation}. Null gives the rate
expected under 2{,}000 within-paper permutations, and
Benjamini--Hochberg adjusted $p$-value across these seven tests, computed separately for
each non-limitation code. Bold marks the code with the higher observed share when
that share is also significant at Adj.\ $p<0.05$.} 
\label{tab:transition_probabilities}
\end{table*}

\section{Transition Frequencies (After Strong Reported Performance)}
\label{app:after_srp}
The \textit{Strong Reported Performance} (SRP) code is one of the most frequent antecedents to limitation disclosures, ranking second only to \textit{Future Work}. 
We therefore examined which limitation codes most often appear immediately after an
SRP statement, again comparing the observed share against the permutation null. 
Table~\ref{tab:srp_precedence} details the limitation codes that are most frequently preceded by SRP statements.

\begin{table}[H]
\centering
\small
\begin{tabular}{@{}l rr rrr@{}}
\toprule
\multirow{2}{*}{\textbf{Limitation Code}} & \multicolumn{2}{c}{\textbf{Occurrences ($n$)}}
& \multicolumn{3}{c}{\textbf{\% Preceded by SRP}} \\
\cmidrule(lr){2-3} \cmidrule(lr){4-6}
& \textbf{SRP $\rightarrow$ Code} & \textbf{All $\rightarrow$ Code}
& \textbf{Obs.} & \textbf{Null} & \textbf{Adj. $p$} \\
\midrule
Empirical Underperformance & 233 & 1,001 & 23.3 & 9.1 & $<$0.001 \\
Methodological Constraints & 645 & 4,941 & 13.1 & 6.9 & $<$0.001 \\
Dependency on Upstream Quality & 264 & 2,042 & 12.9 & 7.2 & $<$0.001 \\
High Time Consumption & 149 & 1,201 & 12.4 & 7.4 & $<$0.001 \\
Model Reasoning and Gen. Deficits & 74 & 614 & 12.1 & 7.2 & $<$0.001 \\
\bottomrule
\end{tabular}
\caption{Limitation categories most frequently preceded by \textit{Strong Reported
Performance} (SRP) statements. Null gives the share percentage expected under 2{,}000
within-paper permutations and Benjamini--Hochberg adjusted $p$-value. All five
observed shares exceed their null, by a factor of 2.6 for \textit{Empirical
Underperformance} and between 1.7 and 1.9 for the remaining four.}
\label{tab:srp_precedence}
\end{table}

\section{Detailed Codebook} \label{app:codebook}
The following tables present the final codebook produced after the full-scale annotation of papers published through 2025.
The limitation codes are ordered by thematic group, where each theme was derived by prompting an LLM to cluster the codes around the question \textit{``Why are these limitations?''} (see Appendix~\ref{app:theme}).
Within each theme, codes are listed in descending order by the number of papers in which they were stated.
The non-limitation codes in Table~\ref{tab:non-lim-codes} follow the same prevalence-based ordering.

\begin{small}
\setlength{\LTleft}{\fill}
\setlength{\LTright}{\fill}

\begin{longtable}[c]{>{\raggedright\arraybackslash}p{0.12\textwidth} >{\raggedright\arraybackslash}p{0.05\textwidth} >{\raggedleft\arraybackslash}p{0.08\textwidth} >{\raggedright\arraybackslash}p{0.35\textwidth} >{\raggedright\arraybackslash}p{0.26\textwidth}}
\caption{Limitation codes from our final codebook (i.e., after the 2025 full-scale run). The year indicates the iteration when the code was added to the codebook, with ``Initial'' denoting codes present in the initial codebook $C_0$. For each code, we provide its definition, total paper count, corpus-wide prevalence (\%), and an exemplary semantic unit.} \\
\toprule
\textbf{Code} & \textbf{Year} & \textbf{Paper \& Prevalence (\%)} & \textbf{Definition} & \textbf{Example Semantic Unit} \\
\midrule
\endfirsthead
\multicolumn{5}{c}{\tablename\ \thetable{} -- continued} \\
\toprule
\textbf{Code} & \textbf{Year} & \textbf{Paper \& Prevalence (\%)} & \textbf{Definition} & \textbf{Example Semantic Unit} \\
\midrule
\endhead
\midrule \multicolumn{5}{r}{\textit{Continued on next page}} \\
\endfoot
\bottomrule
\endlastfoot
\multicolumn{5}{l}{\textit{\textbf{Theme 1: Operating the models requires unsustainable resources}}} \\[2pt]
High Resource Requirements & Initial & 3,285 (20.47\%) & Refers to the requirement for significant processing power, high-performance hardware (e.g., A100 GPUs, TPU pods), or excessive usage of memory (RAM, VRAM, or disk storage), which may prevent the model from being deployed on consumer-grade hardware or edge devices. & \textit{Second, \textbf{due to the limitation of computational resources, we focus on fine-tuning} in this work, leaving applying ROSE to pre-training for future work.} \\ \\[4pt]
High Time Consumption & Initial & 1,453 (9.05\%) & Describes processes that require excessive duration for training, processing, or inference (latency), making the method unsuitable for some applications. & \textit{Due to the complexity of the dynamic programming algorithm, \textbf{it takes almost two days to finish all training epochs} on 1 Tesla V100 GPU (32GB memory).} \\ \\[4pt]
High Financial Cost & Initial & 1,026 (6.39\%) & Refers to prohibitive monetary expenses required to replicate or use the method, such as cloud computing fees, proprietary API costs, or paid datasets. & \textit{First, \textbf{due to budget limit, for the non-instruction following datasets, we only examined our methods with GPT3.5 and Gemini-1.0-Pro} as LLM evaluators.} \\ \\[4pt]
Human Labor and Annotation Bottleneck & 2022 & 862 (5.37\%) & A severe operational constraint where the method relies on heavy, non-scalable manual labor, domain expertise, or human intervention, or suffers from a scarcity, cognitive overload, or lack of quality control of human annotators. & \textit{Finally, \textbf{our approach relies on human feedback on new questions that TeachMe fails to answer or fails to justify indicating significant human efforts.}} \\ \\[4pt]
Scalability Bottleneck & 2022 & 706 (4.40\%) & An algorithmic and architectural limitation where the method experiences severe performance degradation, computational intractability, or exponential resource demands specifically as input lengths, dataset sizes, or model parameters grow. Includes static architectural constraints such as context window limitations or maximum input token lengths, which act as a hard ceiling for processing long-form data. & \textit{Second, \textbf{our graph construction is quadratic in the number of document sentences}, which restricts our method to documents of average length (e.g. 50-70 sentences).} \\ \\[4pt]
High Data Requirement & 2022 & 241 (1.50\%) & Refers to architectural constraints where methods inherently demand massive volumes of training data to function competitively & \textit{One limitation of this work is that the generative model used in this paper \textbf{requires a large amount of training data} and is computationally expensive.} \\ \\[4pt]
Environmental Impact & 2022 & 106 (0.66\%) & The real-world ecological consequences, such as carbon emissions and energy consumption, caused by the computational overhead of training or running large models. & \textit{Thirdly, compared to single-agent approaches, multiple agents require more tokens and time, increasing computational demands \textbf{and environmental impact.}} \\ \\[4pt]
\midrule[0.4pt]
\pagebreak
\multicolumn{5}{l}{\textit{\textbf{Theme 2: Performance breaks down outside narrow experimental constraints}}} \\[2pt]
Scope Limitation & Initial & 10,091 (62.88\%) & Clarifies the deliberate boundaries set by the researchers, detailing what was explicitly excluded from the study (e.g., specific languages, model sizes). & \textit{Despite the strong performance of our two key designs (COCO and iDRO), we \textbf{mainly verify their efficacy from their empirical performance on BEIR tasks}.} \\ \\[4pt]
Methodological Constraints & Initial & 6,382 (39.77\%) & Refers to fundamental restrictions embedded in the design, logic, or mathematical assumptions of the chosen method or architecture that cannot be easily fixed without changing the core approach. Includes: pipeline complexity, preprocessing overhead, training instability, hardware/framework incompatibilities, and the need for task-specific architectural adaptations. & \textit{We use a simple heuristically derived semantic similarity metric, which \textbf{may not fully capture all aspects of semantic similarity between sentence pairs} (Table 3).} \\ \\[4pt]
Generalization Gap & 2022 & 2,414 (15.04\%) & A statistical and domain-transfer limitation where there is an empirical failure or uncertainty regarding a model's ability to maintain performance when transferred across different domains, languages, tasks, context lengths, modalities, cultural contexts, underlying model architectures, hardware platforms, or out-of-distribution datasets (e.g., zero-shot, long-tail, or out-of-vocabulary scenarios). Includes: vulnerability to overfitting on specific datasets or artifacts during training, and sensitivity to noisy, corrupted, or highly variable input data. & \textit{For the latter, we increase our dataset with samples from the CNN/DM dataset, partially mitigating the problem, \textbf{but out-of-domain topics still suffer}.} \\ \\[4pt]
Empirical Underperformance & 2022 & 1,142 (7.12\%) & A quantitative and comparative shortfall where the proposed method explicitly fails to match or exceed the measurable performance metrics of established baselines, state-of-the-art models, human experts, or theoretical upper bounds, or yields only marginal and underwhelming empirical improvements. & \textit{First, \textbf{there is still a performance gap between our method and the vanilla transformer in the language modeling task and CNN/Daily summarization task}.} \\ \\[4pt]
Real-World Deployment Barrier & 2022 & 1,141 (7.11\%) & An ecological validity and operational barrier where the fundamental assumptions, experimental setups, or structured workflows fail to represent the unstructured complexity, noise, and actual user behaviors present in live environments, rendering the method impractical for real-world deployment even if benchmark performance is high. Includes: evaluation datasets or synthetic data failing to faithfully capture the true distribution of live environments. & \textit{Moreover, \textbf{our experiments are conducted on standard benchmark datasets, which may not faithfully represent the noise and complexity of real-world text streams.}} \\ \\[4pt]
Performance Trade-off & Initial & 544 (3.39\%) & Highlights scenarios where an improvement in one metric (e.g., efficiency, speed) results in a direct and unavoidable degradation of another metric (e.g., accuracy, quality). & \textit{Although the decrease of fluency is relatively small compared to the improvement of detoxification, \textbf{MILDecoding does sacrifice language model quality}.} \\ \\[4pt]
Hyperparameter Sensitivity & 2022 & 452 (2.82\%) & A methodological and configuration fragility where a model's performance, stability, or success is highly contingent on precise architectural choices, random seed selection, or extensive manual tuning of mathematical hyperparameters, making the system difficult to optimize or reproduce without exhaustive trial and error. & \textit{Further, the \textbf{need for extensive hyperparameter optimisation, particularly for combining loss functions, can hinder accessibility} for non-expert users.} \\ \\[4pt]
Temporal Degradation & 2022 & 373 (2.32\%) & A vulnerability where a model's effectiveness degrades over time, or a dataset becomes obsolete, due to shifting language patterns, concept drift, or temporal cutoffs. & \textit{This can \textbf{lead to a degraded real-world performance if a system relies exclusively on WebIE for evaluation when the dataset is not updated accordingly}.} \\ \\[4pt]
Prompt Sensitivity & 2022 & 325 (2.03\%) & A linguistic and interaction fragility where a model's output quality, accuracy, or reasoning behavior fluctuates drastically based on minor variations in prompt wording, instructional phrasing, or formatting, demonstrating a lack of robustness to how a task is presented by the user. Includes: a strict dependency on high-quality, representative few-shot examples to function effectively. & \textit{We also note that \textbf{using different prompt templates or changing the phrasing of instruction prompt leads to distinct response behaviors and performance} (Table 5).} \\ \\[4pt]
Literature Coverage Gap & 2024 & 92 (0.57\%) & A methodological and scoping limitation where the authors explicitly acknowledge the potential omission of relevant prior works, recent findings, or valuable contributions due to the rapid expansion, vast volume, or dynamic nature of the scholarly literature in the field & \textit{Although \textbf{we attempted to review the literature comprehensively, the rapid growth of LLM research means some recent preprints might have been missed} (Section 1).} \\ \\[4pt]
\midrule[0.4pt]
\pagebreak
\multicolumn{5}{l}{\textit{\textbf{Theme 3: Systems are bottlenecked by flawed data and external dependencies}}} \\[2pt]
Dependency on Upstream Quality & Initial & 2,549 (15.88\%) & States that the method's performance is strictly bottlenecked by the quality of the underlying base model or preprocessing steps (errors flow downstream). Includes: structural domain mismatches between pre-training and downstream data, inheriting stylistic, structural, or content biases from upstream source data or simulators, and strict dependency on the characteristics of specific source data. & \textit{Our model also \textbf{suffers from the noisy OCR prediction of off-the-shelf object detector}, whose performance will depend highly on the extracted OCR text qualities.} \\ \\[4pt]
Data Scarcity & Initial & 2,019 (12.58\%) & Refers to the overall lack of sufficient data quantity, highlighting situations where there is very little data available or the data belongs to a niche domain, such as low-resource languages or tasks requiring rare examples. Includes: data collection attrition caused by low participant response rates or missing data. & \textit{We clearly realize that \textbf{our dataset size is relatively small compared with other related datasets} due to its unique property of context-sensitiveness.} \\ \\[4pt]
Dataset Bias/Imbalance & Initial & 1,615 (10.06\%) & Acknowledges that the researchers did not have or collect enough data for certain groups, leading to a skewed data distribution, class imbalances, or misrepresentations that contain social/demographic biases (e.g., gender, race). & \textit{\textbf{MTurk workers generally tend to be less religious, more educated, and more likely to be unemployed than the general population} (Goodman et al., 2013).} \\ \\[4pt]
Low Data Quality & 2022 & 1,239 (7.72\%) & Refers to low-quality data containing inaccuracies, typographical errors, wrong labels, formatting issues, or irrelevant content (``garbage'') that confuses the model. Includes: dataset incompleteness, missing annotations or metadata, structural heterogeneity, lack of diversity, and the limitations of relying on unrepresentative synthetic or simulated data Includes: coarse labeling constraints lacking fine-grained granularity, inherent data ambiguity where categories lack clear semantic definitions, artifacts and semantic loss introduced by relying on machine translation & \textit{For instance, \textbf{OntoNotes 5.0 dataset contains 18 coarse-grained entity types while FEW-NERD contains 9 coarse-grained and 66 fine-grained entity types}.} \\ \\[4pt]
Reliance on External Tools or Resources & 2022 & 783 (4.88\%) & Refers to the method's dependency on third-party software, APIs, parsers, or other models to function, which introduces risks of failure if those tools change or go offline. Includes: structural bottlenecks where performance is constrained by the limited coverage or availability of external knowledge bases. & \textit{Moreover, to automatically extract the expected evidence, \textbf{we rely on core NLP tools such as coreference resolution, POS tagger and dependency parsers}.} \\ \\[4pt]
Resource Accessibility Barriers & 2022 & 590 (3.68\%) & A systemic and infrastructural barrier where the broader community is restricted from accessing essential underlying materials due to proprietary data dependencies, closed-source models, copyright constraints, or financial obstacles (e.g., paywalled APIs), hindering the ability to build upon or extend the research. & \textit{Further, many open and close language models are \textbf{trained on content that cannot be acquired or redistributed, and thus could not be included in Dolma.}} \\ \\[4pt]
Dataset Task Mismatch & 2025 & 5 (0.03\%) & A data-centric limitation where the available datasets used for training, fine-tuning, or evaluation do not structurally, semantically, or stylistically align with the target downstream task, specific domain, or intended real-world application. & \textit{\textbf{Those that are available tend to use more general language than scientific medical writing}; this is especially true for the En $\rightarrow$ Ar test set.} \\ \\[4pt]
Sparse Data Sensitivity & 2025 & 3 (0.02\%) & data-centric limitation occurring at the inference or execution stage where the provided individual input data—despite being valid and in-domain—is too brief, fragmentary, or lacks sufficient contextual depth for the model to extract meaningful signals or perform accurate reasoning. & \textit{Sparsity of Belief Signals \textbf{Most users only post about a narrow subset of the belief topics we study, meaning their full belief structure is only partially observable}.} \\ \\[4pt]
\midrule[0.4pt]
\pagebreak
\multicolumn{5}{l}{\textit{\textbf{Theme 4: Scientific claims are difficult to measure and verify}}} \\[2pt]
Measurement Vulnerabilities & 2022 & 957 (5.96\%) & Evaluation flaws where inconsistent experimental setups, lack of comparable baselines, hardware dependencies, missing ablation studies, or saturated benchmarks prevent a fair, comprehensive, and direct comparison with prior state-of-the-art work. Includes: high variance across different data splits, small evaluation scales preventing statistically robust conclusions, lack of expert validation, experimental design biases (e.g., ordering bias, outcome reporting, survivorship bias, or participant observation bias/Hawthorne effect), and proxy or surrogate model discrepancies. & \textit{Although the random selection should ideally represent the whole distribution, \textbf{the performance may vary slightly when comparing to the whole test set.}} \\ \\[4pt]
Subjectivity in Evaluation/Annotation & Initial & 840 (5.23\%) & Refers to the inconsistency, bias, or lack of objectivity inherent in human annotations and assessments due to vague guidelines or personal interpretation. Includes: specific demographic biases introduced by a lack of demographic diversity among human annotators. & \textit{While the benchmark dataset was annotated by human annotators, \textbf{it is important to acknowledge the possibility of annotation errors or inconsistencies.}} \\ \\[4pt]
Theoretical Gap & 2022 & 694 (4.32\%) & An analytical or epistemic limitation where the study identifies empirical success but lacks rigorous mathematical formalization, theoretical guarantees, or fails to isolate the underlying causal mechanisms and confounding variables driving the results. & \textit{\textbf{How to design a unified and more powerful KD method from the perspective of the connection between word- and sequence-level KD still remains unsolved.}} \\ \\[4pt]
Lack of Reliable Evaluation Metrics or Benchmarks & 2022 & 673 (4.19\%) & Points out the absence of established standards, benchmarks, or effective mathematical formulas to measure a specific phenomenon accurately. Includes: structural vulnerabilities where chosen metrics are inadequate, inconsistent, or structurally fail to measure critical qualitative dimensions of model behavior & \textit{However, \textbf{existing benchmarks for low-resource (Han et al., 2018; Gao et al., 2019) are limited to the simple scenario of sentence-level relation classification}.} \\ \\[4pt]
Reliance on Automatic Metrics & Initial & 564 (3.51\%) & Criticizes the dependence on algorithmic scoring systems (e.g., BLEU, ROUGE) which may not correlate with human judgment or semantic understanding. Includes: systematic biases introduced by using LLMs as automated evaluators (e.g., favoring their own generated text, preferring longer responses, or exhibiting instability and inconsistency across multiple evaluation runs). & \textit{Evaluation Metrics Second, we caveat that \textbf{our accuracy metrics (STR-EM and Disambig-F1) only measure the recall of the required information in the long answers.}} \\ \\[4pt]
Data Leakage/Contamination & Initial & 284 (1.77\%) & Mentions the risk of evaluation data (test set) being inadvertently included or ``seen'' during the training phase, leading to invalid or inflated results. & \textit{Lastly, since training data for most models is not publicly available, \textbf{data leakage from the training data corpus could possibly affects our findings}.} \\ \\[4pt]
Lack of Interpretability & Initial & 278 (1.73\%) & Refers to the difficulty in understanding the internal decision-making logic of the model, often described as the ``Black Box'' problem. & \textit{\textbf{The black box nature of these models not only constrains our ability to comprehensively understand their behavior} but also has ethical implications downstream.} \\ \\[4pt]
Reproducibility Gap & 2022 & 181 (1.13\%) & A methodological and transparency barrier where the study lacks sufficient documentation, releases incomplete code, or omits crucial experimental configurations, preventing the broader research community from independently replicating the exact results or validating the original claims. Includes: the inability to accurately reproduce the results of external baselines due to unmaintained repositories or missing implementation details & \textit{\textbf{We use proprietary LLMs like GPT-4 in certain experiments, and those results may be difficult reproduce} if changes to the proprietary services occur.} \\ \\[4pt]
\midrule[0.4pt]
\pagebreak
\multicolumn{5}{l}{\textit{\textbf{Theme 5: The models generate unpredictable and unsafe behaviors}}} \\[2pt]
Model Reasoning and Generation Deficits & 2022 & 673 (4.19\%) & A behavioral and algorithmic limitation where the model exhibits flawed internal logic, fails to capture deep semantic meaning, relies on superficial or spurious cues, suffers from catastrophic forgetting, or produces generic, repetitive, and degenerate outputs. Includes: inherent behavioral instability or non-determinism across inference runs, and susceptibility to reward hacking. & \textit{As the appended adversarial suffix tokens may lack meaning, the \textbf{resulting adversarial prompt found by the attack methods exhibits reduced naturalness}.} \\ \\[4pt]
Bias and Fairness Risks & 2022 & 453 (2.82\%) & An ethical and socio-technical vulnerability where the system or dataset inherits, reproduces, or amplifies historical prejudices, cultural stereotypes, or exhibits structural disparities in performance across different demographic groups. & \textit{For convenience, this work \textbf{uses a standard pronunciation acoustic model from MFA, which may disadvantage those with varying pronunciation preferences}.} \\ \\[4pt]
Model Hallucination/Incoherence & Initial & 375 (2.34\%) & Refers to the generation of output that is factually false, logically nonsensical, repetitive, or fluent but incorrect (common in LLMs). & \textit{At the same time, the \textbf{generated premises could potentially contain fictional information}, and should not be used for training models that learn facts from data.} \\ \\[4pt]
Potential for Misuse & Initial & 246 (1.53\%) & Emphasizes the ethical risks where the technology could be exploited for malicious purposes, fraud, disinformation, or harmful content generation. & \textit{For instance, \textbf{systems might exploit inferred emotional or cognitive states to influence decisions} in commercial, political, or interpersonal contexts.} \\ \\[4pt]
Societal and Ethical Risks & 2022 & 228 (1.42\%) & The unintended real-world harms caused by the deployment of a technology, including legal and liability risks, negative downstream societal impacts, automation bias, deceptive persuasion, and ethical or labor concerns regarding the exploitation of human annotators. & \textit{However, the current model outputs may be factually inconsistent with the input documents, and in such a case \textbf{could contribute to misinformation on the internet}.} \\ \\[4pt]
Privacy and Security Risks & 2022 & 192 (1.20\%) & An ethical and security vulnerability where the system or dataset could inadvertently expose sensitive personal information, user feedback, or be susceptible to adversarial attacks. Includes: vulnerabilities in watermarks, defense mechanisms, paraphrasing attacks, or countermeasures. & \textit{\textbf{A concept has the potential to reveal additional information about the input, which may not be immediately evident} upon inspecting the concept itself.} \\ \\[4pt]
Toxicity Risks & 2022 & 128 (0.80\%) & A behavioral and content-safety vulnerability where the system could inadvertently generate or be provoked into producing abusive language, hate speech, or other violent, offensive, and inappropriate content. & \textit{Moreover, the \textbf{training datasets we used contain violence, abuse, and biased content} that can be upsetting or offensive to particular groups of people.} \\ \\[4pt]
\end{longtable}

\clearpage

\setlength{\LTleft}{\fill}
\setlength{\LTright}{\fill}

\begin{longtable}[c]{>{\raggedright\arraybackslash}p{0.12\textwidth} >{\raggedright\arraybackslash}p{0.05\textwidth} >{\raggedleft\arraybackslash}p{0.08\textwidth} >{\raggedright\arraybackslash}p{0.35\textwidth} >{\raggedright\arraybackslash}p{0.26\textwidth}}
\caption{Non-Limitation (NL) codes from our final codebook (i.e., after the 2025 full-scale run). The year indicates the iteration when the code was added to the codebook, with ``Initial'' denoting codes present in the initial codebook $C_0$. For each code, we provide its definition, total paper count, corpus-wide prevalence (\%), and an exemplary semantic unit.} \label{tab:non-lim-codes} \\
\toprule
\textbf{Code} & \textbf{Year} & \textbf{Paper \& Prevalence (\%)} & \textbf{Definition} & \textbf{Example Semantic Unit} \\
\midrule
\endfirsthead
\multicolumn{5}{c}{\tablename\ \thetable{} -- continued} \\
\toprule
\textbf{Code} & \textbf{Year} & \textbf{Paper \& Prevalence (\%)} & \textbf{Definition} & \textbf{Example Semantic Unit} \\
\midrule
\endhead
\midrule \multicolumn{5}{r}{\textit{Continued on next page}} \\
\endfoot
\bottomrule
\endlastfoot
NL: Future Work & Initial & 12,041 (75.04\%) & Suggests potential improvements or delegates specific unaddressed features, unanswered questions, or planned extensions to be explored in subsequent studies. Includes: theoretical mitigations proposed by the authors. & \textit{\textbf{There are also ways to incorporate model generated labelling methods for more robust semi-supervision into our framework that we leave to future work.}} \\ \\[4pt]
NL: Contextual Justification & Initial & 6,561 (40.89\%) & Refers to explanations that provide external background, common field practices, or real-world constraints to defend and justify a limitation. & \textit{This is largely due to \textbf{copyright issues involved with using more modern texts, and because of this, our models are inherently biased to the language used in older texts.}} \\ \\[4pt]
NL: Method Details & Initial & 5,016 (31.26\%) & Refers to factual descriptions of the proposed method which are typically stated to explain the root cause of a specific limitation. & \textit{Intuitively, the \textbf{connection between nodes in the input graph can influence the encoding of x by guiding what to extract} from x in order to generate y.} \\ \\[4pt]
NL: Strong Reported Performance & Initial & 4,668 (29.09\%) & Refers to statements explicitly highlighting the model's empirical success, effectiveness, or superior benchmark results. & \textit{Although \textbf{MABEL shows exciting performance across an extensive range of evaluation settings}, these results should not be construed as a complete erasure of bias.} \\ \\[4pt]
NL: Theoretical Projection & 2022 & 2,600 (16.20\%) & Optimistic or theoretical guesses, as well as explicit uncertainties, regarding how a method might perform in untested scenarios, unseen domains, or with future parameter and scale adjustments. & \textit{\textbf{We expect improvements in classification performance when applied to multiclass or multilabel classification settings, but we have not confirmed this.}} \\ \\[4pt]
NL: Conducted Mitigation & Initial & 2,366 (14.74\%) & Describes proactive steps, alternative methods, or workarounds that the researchers have already implemented within the study to alleviate a limitation. Includes: failed mitigation attempts. & \textit{\textbf{To reduce the impact of LM bias, we generate hundreds of thousands of test cases to increase the chance we obtain test cases for a given sub-category.}} \\ \\[4pt]
NL: Method Strength & Initial & 2,021 (12.59\%) & Highlights the theoretical, structural, or conceptual advantages of the proposed approach, distinct from empirical performance. & \textit{In this work, \textbf{we propose a method to draw structured robust early-bird tickets, which can be used as an efficient alternative to adversarial training.}} \\ \\[4pt]
NL: Anticipated Impact & Initial & 1,508 (9.40\%) & Expresses the authors' hopes, expectations, or the broader potential influence that their findings might have on the research community or industry. & \textit{\textbf{We hope our findings can inform potential avenues of improvement on data augmentation for NER and inspire the further work} in this research direction.} \\ \\[4pt]
NL: Recommendations & 2022 & 1,106 (6.89\%) & Outward-facing directives advising the broader research or practitioner community to adopt specific operational practices, metrics, strategies, or to contribute to specific research directions. & \textit{In the long-run, \textbf{the community should build more elaborate coherence measures, to build a more complete picture of model capabilities and limitations.}} \\ \\[4pt]
NL: Authorial Disclaimers & 2022 & 839 (5.23\%) & Explicit epistemic boundaries and warnings drawn by authors to prevent readers from overclaiming, misinterpreting, or prematurely deploying the empirical results in real-world or high-stakes applications. & \textit{\textbf{Regardless, we believe our models should be utilized with caution and approaches to mitigating social risks, biases, and toxicities should be carefully applied.}} \\ \\[4pt]
\end{longtable}

\end{small}
\clearpage
\section{Codebook Evolution Summary}
\label{app:codebook_evolution}
Through the iterative codebook update process, we compile the empirical statistics of codebook evolution in Table~\ref{tab:appendix_codebook_updates}, reflecting the final human-approved decisions in response to LLM-generated recommendations. Specifically, for each iteration, LLM-suggested actions (i.e., merging, expanding, adding new codes, and rejecting) were manually reviewed and confirmed or altered by humans in the loop to update the codebook. Detailed codebook updates by iteration can be found in Table~\ref{tab:codebook_updates_details}.

\begin{table}[h]
\centering
\small
\begin{tabular}{lcccc}
\toprule
 & \textbf{2020--2022} & \textbf{2023} & \textbf{2024} & \textbf{2025} \\
\midrule
No. of unique new codes detected & 104 & 206 & 200 & 228 \\
No. of clusters formed           & 105 & 104 & 143 & 74 \\
\quad -- \textsc{Add}                   & 22  & 0   & 1   & 2  \\
\quad -- \textsc{Expand}                   & 6   & 9   & 4   & 8  \\
\quad -- \textsc{Merge}                    & 75  & 95  & 138 & 64 \\
\quad -- \textsc{Reject}                   & 0   & 0   & 0   & 0  \\
\bottomrule
\end{tabular}
\caption{Summary of codebook update actions across the four iterations, representing the final decisions made by humans in the loop.}
\label{tab:appendix_codebook_updates}
\end{table}

\bigskip

\begin{table*}[h]
\centering
\small
\begin{tabular}{llp{10.4cm}}
\toprule
\textbf{Period} & \textbf{Action} & \textbf{Affected Codes} \\
\midrule
\textbf{2020--2022} & \textbf{ADD NEW (22)} & 
\textit{Bias and Fairness Risks} ~$\bullet$~
\textit{Toxicity Risks} ~$\bullet$~
\textit{Empirical Underperformance} ~$\bullet$~
\textit{High Data Requirement} ~$\bullet$~
\textit{Environmental Impact} ~$\bullet$~
\textit{Societal and Ethical Risks} ~$\bullet$~
\textit{Measurement Vulnerabilities} ~$\bullet$~
\textit{Generalization Gap} ~$\bullet$~
\textit{Scalability Bottleneck} ~$\bullet$~
\textit{Human Labor and Annotation Bottleneck} ~$\bullet$~
\textit{Model Reasoning and Generation Deficits} ~$\bullet$~
\textit{NL: Recommendations} ~$\bullet$~
\textit{NL: Theoretical Projection} ~$\bullet$~
\textit{NL: Authorial Disclaimers} ~$\bullet$~
\textit{Privacy and Security Risks} ~$\bullet$~
\textit{Real-World Deployment Barrier} ~$\bullet$~
\textit{Resource Accessibility Barriers} ~$\bullet$~
\textit{Reproducibility Gap} ~$\bullet$~
\textit{Hyperparameter Sensitivity} ~$\bullet$~
\textit{Prompt Sensitivity} ~$\bullet$~
\textit{Temporal Degradation} ~$\bullet$~
\textit{Theoretical Gap} \\
\cmidrule{2-3}
 & \textbf{EXPAND (6)} & 
\textit{Low Data Quality} ~$\bullet$~
\textit{Lack of Reliable Evaluation Metrics or Benchmarks} ~$\bullet$~
\textit{Methodological Constraints} ~$\bullet$~
\textit{NL: Future Work} ~$\bullet$~
\textit{NL: Conducted Mitigation} ~$\bullet$~
\textit{Reliance on External Tools or Resources} \\
\midrule
\textbf{2023} & \textbf{ADD NEW (0)} & None \\
\cmidrule{2-3}
 & \textbf{EXPAND (6 unique)} & 
\textit{Empirical Underperformance} ~$\bullet$~
\textit{Generalization Gap} ~$\bullet$~
\textit{Reliance on Automatic Metrics} ~$\bullet$~
\textit{Measurement Vulnerabilities} ~$\bullet$~
\textit{Dependency on Upstream Quality} ~$\bullet$~
\textit{Subjectivity in Evaluation/Annotation} \\
\midrule
\textbf{2024} & \textbf{ADD NEW (1)} & 
\textit{Literature Coverage Gap} \\
\cmidrule{2-3}
 & \textbf{EXPAND (4)} & 
\textit{Model Reasoning and Generation Deficits} ~$\bullet$~
\textit{Measurement Vulnerabilities} ~$\bullet$~
\textit{Reliance on Automatic Metrics} ~$\bullet$~
\textit{Societal and Ethical Risks} \\
\midrule
\textbf{2025} & \textbf{ADD NEW (2)} & 
\textit{Sparse Data Sensitivity} ~$\bullet$~
\textit{Dataset Task Mismatch} \\
\cmidrule{2-3}
 & \textbf{EXPAND (8)} & 
\textit{Scalability Bottleneck} ~$\bullet$~
\textit{Data Scarcity} ~$\bullet$~
\textit{Generalization Gap} ~$\bullet$~
\textit{Low Data Quality} ~$\bullet$~
\textit{Privacy and Security Risks} ~$\bullet$~
\textit{Prompt Sensitivity} ~$\bullet$~
\textit{Real-World Deployment Barrier} ~$\bullet$~
\textit{Reproducibility Gap} \\
\bottomrule
\end{tabular}
\caption{Detailed codebook updates: specific codes added and expanded by iteration. The ``NL:'' prefix indicates a non-limitation code.}
\label{tab:codebook_updates_details}
\end{table*}

\newpage
\section{Implementation Details}
\subsection{Affiliation Extraction}
\label{appendix:affiliation_ext}
The affiliations were extracted using a custom extraction pipeline. First, the raw PDF files were retrieved from the ACL Anthology Python library, and their first pages were parsed using PyMuPDF library\footnote{\url{https://github.com/pymupdf/PyMuPDF}}. Author names and affiliations were then identified from the parsed text using regular expressions. If this parser failed to detect affiliations, a fallback parser using Docling’s OCR engine was used to re-parse and detect affiliations again. The remaining papers without any affiliations were then manually corrected by a human checker. To evaluate the pipeline's performance, we manually examined a random sample of 50 papers, achieving a 96\% agreement rate with a human checker.

\subsection{Affiliation Type Assignment}
\label{appendix:affiliation_label}
To assign papers into the three affiliation groups, we first used an LLM to classify each unique extracted affiliation string into one of four granular types: (1) Large Company (corporations listed in the Forbes Global 2000), (2) Non-large Company (academic institutions and non-large companies), (3) Mixed Affiliation (extracted text containing both large and non-large entities due to parsing artifacts), and (4) Non-affiliation (parsing artifacts such as location, email, publication year, or symbols). Classification was performed using batch inference with a batch size of 30. After assigning these types to unique affiliations, we mapped each paper to a final group based on the following rules (excluding "Non-affiliation" entries):

\begin{itemize}
    \item \textbf{Large-company Group}: If all affiliations in the paper are Large Company.
    \item \textbf{Non-large Company Group}: If all affiliations in the paper are Non-large Company.
    \item \textbf{Mixed affiliation Group}: The paper contains a combination of Large Company and Non-large Company types, or includes a Mixed Affiliation type affiliation.
\end{itemize}

The prompt used for affiliation classification is shown in Prompt \mbox{~\ref{box:llm-affiliation-prompt}}. To evaluate assignment quality, a human annotator randomly sampled 80 unique affiliations and verified agreement with the model outputs. The LLM classification method achieved a 90\% agreement rate with the human annotator.

\begin{promptbox}{Prompt~{\hypersetup{linkcolor=white}\ref{appendix:affiliation_label}}\quad LLM-based affiliation labeling.}
\label{box:llm-affiliation-prompt}
\textbf{System: }You are an expert academic research metadata classifier.
Your task is to classify institution/organization affiliation strings from AI/NLP research papers into one of four categories:

1. "Large Company": Pure commercial enterprises listed in the Forbes Global 2000 top companies list, or their direct research labs, AI divisions, and acquired subsidiaries (e.g. Google, Alphabet, Google Brain, Google DeepMind, Microsoft, Meta FAIR, Amazon AWS, IBM Research, Apple, Nvidia, Tencent, Alibaba, Samsung, etc.).

2. "Non-Large Company": Pure universities, academic institutes, colleges, government labs (e.g. NIST, AIST, CNRS), non-profit research institutes (e.g. Allen Institute for AI / AI2), independent research labs, or small startups/companies NOT listed in Forbes Global 2000.

3. "Mixed": Single unseparated affiliation string that contains BOTH a Large Company AND a Non-Large Company / Academic Institution joined together (e.g. "University of California, Los Angeles, CA, USA Amazon Alexa AI, Manhattan Beach, CA, USA" or "Stanford University / Google Research"). Exclude the string that contain two or more Non-Large Company affiliations.

4. "Non-Affiliation": False positives, non-institutional text, personal names, location, email addresses, web URLs, page numbers, license text, or OCR artifacts accidentally extracted as affiliation strings.

Return the classification results strictly matching the requested schema.

Below is the official reference list of Forbes Global 2000 companies for exact reference: \{ COMPANY\_LIST \} \newline
\tcbline
\textbf{Human} \newline
Classify each of the following research paper affiliation strings as either "Large Company", "Non-Large Company", "Mixed", or "Non-Affiliation": \texttt{\{input\_text\}}
\end{promptbox}

\subsection{Semantic Unit Segmentation Prompt}
This prompt was used as the system instruction for the LLM segmenter (Gemini 2.5 Flash). The human message contains the raw Limitations section of each paper. To maximize inference speed during the segmentation process, the model's thinking configuration budget and temperature was explicitly set to zero. We employed this LLM-driven approach rather than conventional sentence splitting to ensure that the semantic content remains fully intact and free from improper fragmentation. The prompt is shown in Prompt~\ref{box:prompt-segmentation}.

\begin{promptbox}{Prompt~{\hypersetup{linkcolor=white}\ref{box:prompt-segmentation}}\quad Semantic unit segmentation ($M_{\text{seg}}$).}
\label{box:prompt-segmentation}
\textbf{System} \\[3pt]
Split the text below into semantically distinct chunks (sentences or meaningful clauses). \newline
Output format: JSON list of strings. \newline
CRITICAL: Preserve the original text EXACTLY (verbatim). Do not modify, omit, or summarize anything.
\tcbline
\textbf{Human} \\[3pt]
\texttt{\{input\_text\}}
\end{promptbox}
\subsection{Hybrid Qualitative Coding Prompt}
\label{sec:coding-prompt}
The main sentence-level annotation prompt is shown in Prompt~\ref{box:prompt-hybrid-coding}. The placeholder \texttt{\{codebook\}} is filled with the current codebook version; \texttt{\{examples\}} is filled with few-shot examples. To ensure deterministic outputs, the model's temperature was set to 0. Also, to leverage deep reasoning capabilities during qualitative coding, the model was set to operate in high-thinking mode. During pipeline development, we evaluated both individual unit-by-unit annotation and batch-wise processing, finding that the resulting outputs exhibited no significant difference in annotation quality. Consequently, to optimize inference speed, maximize context efficiency, and prevent model confusion caused by overly congested prompts, we settled on a target size of approximately 100 semantic units per batch. Crucially, this boundary is configured to dynamically extend beyond the 100-unit threshold to ensure that the final paper included in a batch has all of its constituent semantic units processed entirely together, thereby preventing document fragmentation across batches.

\begin{promptbox}{Prompt~{\hypersetup{linkcolor=white}\ref{box:prompt-hybrid-coding}}\quad Hybrid qualitative coding ($M_{\text{code}}$).
}
\label{box:prompt-hybrid-coding}
\textbf{System} \\[3pt]
You are an expert qualitative researcher specializing in Natural Language Processing (NLP) literature. Your task is to perform exhaustive sentence-level Hybrid Thematic Analysis on academic texts, identifying and coding limitations, justifications, and future works based on a specific codebook. \newline
\textbf{THE CODEBOOK} \newline
\{codebook\} \newline
\par\smallskip
1. NESTED BATCH PROCESSING: You will receive sentences wrapped within \texttt{$<$section paper\_id=''...''$>$} tags. You MUST read the ENTIRE section first to grasp the overarching narrative, limitations, and causal links specific to that paper. Then, perform your analysis and output codes for EACH \texttt{$<$sentence$>$} sequentially. Do not bleed context from one \texttt{$<$section$>$} into another. \newline
2. CROSS-SENTENCE CONTINUITY (AVOID REDUNDANCY): Track the narrative flow within the section. Do NOT redundantly assign the exact same limitation flaw code to consecutive sentences unless a fundamentally new aspect is introduced. If a subsequent sentence merely continues describing a previously coded limitation without adding new functional consequences or justifications, output exactly \texttt{Code: None}. \newline
3. PARSIMONY PRINCIPLE (NO LABEL SPRAWL): Dissect the sentence to extract distinct limitations, but DO NOT over-segment. Choose ONLY the most core, primary issues. Limit your output to a maximum of 1 to 2 distinct codes per sentence. Do NOT over-predict. \newline
4. AGGRESSIVE NOISE FILTER: If the sentence is just a section header (e.g., "Limitations"), a neutral background preamble, standard dataset statistics without critique, or a transitional filler phrase, output exactly \texttt{Code: None}. Do not force a code on a sentence that lacks analytical weight. \newline
5. STRICT TRIGGERS \& ROOT CAUSE: \newline
    - Do not trigger codes based purely on isolated keywords. Analyze the underlying fundamental bottleneck. \newline
    - Optimism about the future is strictly 'Non-Limitation: Future Work', NOT Conducted Mitigation or Anticipated Impact. \newline
    - Contextual Justification requires a defensive argument, not just a causal explanation. \newline
6. CODE ASSIGNMENT \& INDUCTIVE FREEDOM: \newline
    - The Codebook is your foundation, not your ceiling. You are heavily encouraged to practice "Grounded Theory" by generating New Codes. \newline
    - IN-VIVO FIRST: Always extract the exact functional noun phrase from the text describing the flaw (e.g., 'False-Negative Annotations', 'Excessive Decoding Latency'). \newline
    - DOUBLE-LAYER MANDATE: You possess the freedom to pair a granular New Code alongside ANY broad predefined limitation code (MC, HRR, Data Scarcity) to capture specific technical anatomical details. \newline
    - Avoid hallucinating generic terms. Your New Codes MUST be directly synthesized from the authors' exact terminology (Text-Derived Naming). \newline
\par\smallskip
\textbf{OUTPUT FORMAT EXPECTED} \newline
You MUST strictly follow this exact plain-text structure for EVERY sentence. Do not add conversational filler. \newline
\par\smallskip
--- Sentence ID: [ID] --- \newline
Analysis [N]: \newline
- Focus: "[Exact quote from the sentence]" \newline
- Reasoning: "[Your step-by-step reasoning comparing the focus to the codebook]" \newline
- Code: "[Exact Code Name from Codebook, a completely NEW Code Name, or None]" \newline
\par\smallskip
\textbf{FEW-SHOT EXAMPLES} \newline
\{examples\} \newline
\par\smallskip
\textbf{REMINDER BEFORE YOU START} \newline
- Read the entire \texttt{$<$section$>$} block for context before analyzing its sentences. \newline
- Analyze EVERY \texttt{$<$sentence$>$} provided in the batch. \newline
- Output the \texttt{--- Sentence ID: [ID] ---} header before the analyses for each sentence. \newline
- Ensure exhaustiveness (Analysis 1, Analysis 2...) if a sentence compounds fundamentally independent issues, but strictly adhere to the Parsimony Principle. \newline
\par\smallskip
\textbf{CURRENT INPUT BATCH}
\tcbline
\textbf{Human} \\[3pt]
\{input\_batch\}
\end{promptbox}
\subsection{New Code Consolidation Prompt}
This prompt (\ref{box:prompt-consolidation}) was run after each annotation iteration on newly generated codes. 
As with the previous prompt,
the model was configured with a temperature of 0 and operated in high-thinking mode. The placeholder \texttt{\{NEW\_CODES\}} is a structured list of new code labels and their evidence semantic units.

\begin{promptbox}{Prompt~{\hypersetup{linkcolor=white}\ref{box:prompt-consolidation}}\quad New code consolidation ($M_{\text{consolidate}}$).}
\label{box:prompt-consolidation}
\textbf{System} \\[3pt]
You are an expert qualitative researcher and NLP scientist. Your task is to perform "Code Consolidation" on a list of newly generated codes (Emergent Codes) extracted from multiple separate analysis batches. \newline
\par\smallskip
Because these codes were generated independently, many are semantic synonyms or slight lexical variations of the exact same specific NLP phenomenon. Your goal is to merge these redundant codes into unified **Master Codes** at the specific code-level, while filtering out weak or idiosyncratic outliers as noise. \newline
\par\smallskip
\textbf{PRINCIPLES OF CODE CONSOLIDATION} \newline
1. SEMANTIC CONSOLIDATION (STAY AT CODE LEVEL): Merge codes that describe the *exact same* specific bottleneck, phenomenon, or structural flaw. Select the most precise, standard academic ML/NLP term for the Master Code. \newline
   - CRITICAL: Do NOT over-generalize into broad umbrella themes. (e.g., You can merge "Prompt Instability" and "Sensitivity to Prompts" into \texttt{Prompt Sensitivity}, but do NOT generalize them all the way up to "Evaluation Issues" or "Lack of Robustness"). Keep the granularity fine-grained. \newline
2. NOISE \& OUTLIER FILTERING: Evaluate the strength and generalizability of the evidence. If an input code represents a hyper-specific edge case, has extremely weak/insufficient evidence, or does not represent a meaningful scientific limitation/justification, DO NOT create a Master Code for it. Instead, merge it into a special category named exactly \texttt{Noise / Idiosyncratic Outliers}. \newline
3. EVIDENCE-BASED MERGING: Do not merge codes based purely on similar names. Read the \texttt{Original Evidence} and its \texttt{Context} carefully to ensure they are truly identical phenomena before merging. \newline
4. PREFIX PRESERVATION: If you are merging emergent codes that represent positive aspects, justifications, or strengths, the resulting Master Code MUST retain the strict prefix \texttt{Non-Limitation: }. \newline
\par\smallskip
\textbf{INPUT FORMAT} \newline
You will receive a list of emergent codes. Each entry contains: \newline
- Code Name: [The original emergent name] \newline
- Original Evidence: [Bullet points of raw sentences and context extracted from papers] \newline
\par\smallskip
\textbf{OUTPUT FORMAT EXPECTED} \newline
Output a structured list of your finalized Master Codes. Every original code provided in the input MUST be accounted for (either merged into a specific Master Code, or dumped into \texttt{Noise / Idiosyncratic Outliers}). \newline
\par\smallskip
--- Master Code: [Highly Specific NLP Term OR "Noise / Idiosyncratic Outliers"] --- \newline
- Definition: [A clear, 1-2 sentence academic definition. (Skip this field if the code is Noise)] \newline
- Merged Original Codes: [List the exact names of all input codes that were grouped into this Master Code.] \newline
- Justification: [Briefly explain WHY these codes were merged based on their evidence, or why they were deemed Noise.] \newline
\par\smallskip
\textbf{CURRENT INPUT OF EMERGENT CODES} \newline
\{NEW\_CODES\}
\end{promptbox}
\subsection{Codebook Update Recommendation Prompt}
This prompt suggests an appropriate action for each consolidated cluster: \textsc{merge}, \textsc{expand}, \textsc{add new}, or \textsc{reject} (Prompt~\ref{box:prompt-recommendation}). 
As with the previous prompts, 
the model was configured with a temperature of 0 and operated in high-thinking mode. Within this prompt, the \texttt{\{CURRENT\_CODEBOOK\}} placeholder was populated with the active version of the codebook slated for refinement, while the \texttt{\{LLM\_CLUSTER\_DATAFRAME\}} placeholder inputted the newly consolidated codes and their associated semantic groups derived from the preceding step.

\begin{promptbox}{Prompt~{\hypersetup{linkcolor=white}\ref{box:prompt-recommendation}}\quad Codebook update recommendation ($M_{\text{rec}}$).}
\label{box:prompt-recommendation}
\textbf{System} \\[3pt]
You are an expert qualitative researcher and a methodologically rigorous auditor specializing in NLP literature. Your task is to analyze semantic clusters of newly generated limitation codes (synthesized by an LLM) and recommend data-driven updates to our existing Codebook. \newline
\par\smallskip
Crucially, your output will be used to automatically re-assign the underlying granular codes to their final, official labels. \newline
\par\smallskip
\textbf{PRINCIPLES OF CODEBOOK MANAGEMENT (STRICT GATES)} \newline
You must adhere to the principle of "Parsimony" (keeping the codebook as concise as possible). Do NOT suggest adding a new code unless the evidence is overwhelming. \newline
1. The Synonym Trap: If a proposed cluster is merely a synonym, subset, or specific instance of an existing code, DO NOT add a new code. \newline
2. The Artifact Trap: Reject clusters that represent generic LLM meta-language, overly broad concepts (e.g., "General Limitation"), or noise. \newline
3. The Threshold for New Codes: A cluster warrants a completely NEW code ONLY IF it represents a fundamentally distinct concept, has a clear academic definition, and has a significant 'Total Evidence'/'Total Sources' count that cannot be absorbed by existing categories. \newline
4. Non-Limitation Strictness (The Broad Bucket Rule): For clusters related to mitigations, future plans, strengths, or justifications, strongly prefer to [MERGE] them into the existing broad "Non-Limitation: ..." categories. \newline
5. Non-Limitation Naming Convention: If you absolutely MUST apply [ADD NEW] or [EXPAND] to a non-limitation concept, the new or revised code name MUST strictly begin with the prefix "Non-Limitation: " (e.g., Non-Limitation: Open-Source Release). \newline
6. Cross-Cluster Deduplication: BEFORE assigning actions, review the ENTIRE batch of current semantic clusters. If multiple proposed clusters overlap with each other, consolidate them. Choose ONE unifying name to be the [ADD NEW], and assign [MERGE] to the redundant cluster(s), mapping them to that newly unified name. \newline
7. Structured Expansion Principle (PRESERVE \& INTEGRATE): When an \texttt{[EXPAND]} action is triggered, apply the following rules **in order**: \newline
  1. Deduplicate first. If the new nuance is already covered — explicitly or implicitly — by the existing definition, make **no changes**. \newline
  2. Merge elegantly. If the nuance is genuinely new, integrate it using one of these methods: \newline
    - Weave it naturally into existing example lists or phrasing \newline
    - Append it as a concise bullet under an \texttt{Includes:} or \texttt{Examples:} clause at the end of the definition \newline
  3. Preserve the core. Never delete, truncate, rewrite, or overwrite original sentences. The foundational meaning must remain intact and unchanged. \newline
8. Consolidated Expansion: If multiple proposed clusters in the current batch trigger an [EXPAND] action for the EXACT SAME existing code, you must synthesize ALL of their nuances into a single, unified 'Revised Definition'. Output this same unified definition for every cluster that expands that specific code. \newline
\par\smallskip
\textbf{EXISTING CODEBOOK} \newline
\{CURRENT\_CODEBOOK\} \newline
\par\smallskip
\textbf{INSTRUCTIONS FOR ACTIONS \& TRACEBACK MAPPING} \newline
I will provide you with a dataframe of Semantic Clusters. For each Cluster, evaluate its proposed 'code' name, 'definition', and the list of 'original\_codes' it contains. Choose ONE of the following four actions: \newline
\par\smallskip
- [MERGE]: The cluster is redundant. It aligns perfectly with an existing code OR it overlaps with another new cluster in this current batch. \newline
  -> Traceback Action: All 'original\_codes' in this cluster must be re-assigned to the EXACT NAME of the existing code, OR to the unified New Code Name you approved for the overlapping cluster. \newline
- [EXPAND]: The cluster belongs to an existing code, but highlights a specific nuance that the current definition misses. \newline
  -> Traceback Action: All 'original\_codes' must be re-assigned to the existing code. You must provide an ADDITIVE updated definition that preserves the original meaning while appending the new nuance(s). \newline
- [ADD NEW]: The cluster is highly prevalent, academically distinct, represents a true blind spot, AND has been deduplicated against other clusters in this batch. \newline
  -> Traceback Action: Approve the proposed code name (or refine it to be more formal) and its formal definition. All 'original\_codes' will be assigned to this New Label. \newline
- [REJECT]: The cluster is noisy, vague, or lacks coherent meaning. \newline
  -> Traceback Action: Do not map these codes. \newline
\par\smallskip
\textbf{OUTPUT FORMAT EXPECTED} \newline
Analyze each cluster systematically using this exact format to allow for automated parsing: \newline
\par\smallskip
--- Cluster ID: [Index/ID] | Proposed Name: [code] --- \newline
1. Evaluation: \newline
   - Prevalence: [Mention Total Evidence and Total Sources] \newline
   - Concept Mapping: [Compare against the Existing Codebook AND other clusters in this batch.] \newline
2. Justification: [Argue strictly WHY this needs a change, WHY it perfectly matches an existing code, or WHY it is being merged/expanded.] \newline
3. Action: [MERGE / EXPAND / ADD NEW / REJECT] \newline
4. Traceback Mapping \& Recommendation: \newline
   - Target Label: [The EXACT existing code name, the approved New Code Name, or the unified cross-cluster name. If REJECT, write "NONE".] \newline
   - Revised Definition: [If EXPAND, provide the ADDITIVE new definition preserving the original text. If ADD NEW, provide the formal definition. If MERGE/REJECT, write "N/A". Ensure consistency if expanding the same target label multiple times.] \newline
\par\smallskip
\textbf{CURRENT SEMANTIC CLUSTERS} \newline
\{LLM\_CLUSTER\_DATAFRAME\}
\end{promptbox}

\subsection{Second-Level Intra-Code Clustering Prompt}
This prompt explores sub-codes within a single parent code by clustering its evidence sentences (Prompt~\ref{box:prompt-clustering}). Within this prompt, the \texttt{\{target\_code\_name\}} placeholder specifies the particular parent code currently under analysis, while the \texttt{\{input\_json\}} placeholder delivers the structured input data comprising all semantic units associated with that target code, thereby enabling the model to investigate and uncover more granular, fine-grained sub-structures within its content.
To guarantee strict output determinism and leverage deep structural analysis, the model was configured with a temperature of 0 and operated in high-thinking mode. 

\begin{promptbox}{Prompt~{\hypersetup{linkcolor=white}\ref{box:prompt-clustering}}\quad Second-level intra-code clustering.}
\label{box:prompt-clustering}
\textbf{System} \\[3pt]
You are an expert Qualitative Researcher conducting a 2nd-level Thematic Analysis. \newline
You are analyzing a JSON array of sentences that share the overarching parent code: "\{target\_code\_name\}". \newline
\par\smallskip
INSTRUCTIONS \& CHAIN OF THOUGHT: \newline
Step 1: Concept Extraction --- identify recurring mechanisms, boundaries, or subjects being restricted. \newline
Step 2: Taxonomy Generation --- group into distinct, mutually exclusive clusters based on core mechanism. \newline
Step 3: In-Vivo Naming --- precise academic noun phrase from the authors' own terminology. No generic terms. \newline
Step 4: Anchor Selection --- EXACTLY 5 representative sentences per cluster (or all if fewer than 5). \newline
Step 5: Keyword Extraction --- 3--5 high-signal keywords/n-grams for future trend tracking. \newline
\par\smallskip
INPUT DATA: \newline
\{input\_json\} \newline
\par\smallskip
OUTPUT FORMAT (JSON only, no markdown fences): \newline
\{"Clusters": [\{"cluster\_name": "...", "inclusion\_criteria": "...", "reasoning": "...", \newline
 "representative\_sentences": ["..."], "keywords": ["..."]\}]\}
\end{promptbox}

\subsection{Question-Based Theme Searching Prompt} \label{app:theme}
This prompt groups the finalized codes into overarching themes that directly address a given research question (Prompt~\ref{box:prompt-theme-generation}). The \texttt{\{question\}} placeholder specifies the target question of interest for theme searching, while the \texttt{\{code\}} placeholder delivers the comprehensive details of each code, including its name, definition, semantic unit examples, and overall prevalence.
Within our current pipeline, these searched themes were utilized primarily to systematically sort and structure the codebook for enhanced readability and presentation. To facilitate deep qualitative synthesis, the model operated in high-thinking mode with a default temperature of 0. 
Notably, this temperature can be increased if generating a more diverse set of themes is desired. 
\begin{promptbox}{Prompt~{\hypersetup{linkcolor=white}\ref{box:prompt-theme-generation}}\quad RQ-driven theme generation.}
\label{box:prompt-theme-generation}
\textbf{System} \\[3pt]
You are an expert qualitative researcher and theorist specializing in Natural Language Processing (NLP). \newline
You have just completed the coding phase of a Thematic Analysis. \newline
Your task is to synthesize the final Enriched Codebook into overarching Themes that specifically answer a defined Research Question. \newline
\par\smallskip
\textbf{RESEARCH QUESTION} \newline
\{question\} \newline
\par\smallskip
\textbf{PRINCIPLES OF RQ-DRIVEN THEME GENERATION} \newline
1. Answer the RQ: Every theme MUST directly answer the Research Question — not just categorize codes. \newline
2. Latent Meaning over Semantic Similarity: Group codes based on how they collectively answer the RQ, even if they don't share keywords. \newline
3. Weight vs. Significance: Use Prevalence to gauge gravity. High-count codes often form the core of major themes. \newline
4. Exclusivity: A code should primarily belong to only ONE theme to maintain a clear narrative. \newline
5. Simple, Scannable Theme Names: Each theme name must be immediately understandable on first read. \newline
   - Use plain language — no jargon, no academic abstractions. \newline
   - Prefer short noun phrases or a single clear sentence (under 10 words). \newline
   - A reader unfamiliar with the study should grasp the theme's meaning instantly. \newline
\par\smallskip
\textbf{ENRICHED CODEBOOK} \newline
\{code\} \newline
\par\smallskip
\textbf{INSTRUCTIONS} \newline
Group ALL provided codes into exactly 3 to 5 distinct Themes. \newline
For each Theme, define its "Core Answer to the RQ". \newline
\par\smallskip
\textbf{OUTPUT FORMAT} \newline
--- Theme [N]: [Theme Name — a narrative phrase that answers the RQ] --- \newline
- Core Answer to the RQ: [2-3 sentences explaining HOW this theme answers the RQ.] \newline
- Constituent Codes: \newline
  * [Code Name] (Prevalence: [N papers]) - [1 sentence on how this code supports the theme.]
\tcbline
\textbf{Human} \\[3pt]
\texttt{\{enriched\_codebook\}}
\end{promptbox}

\section{The Use of AI Assistants}
In this research, we used Large Language Models (LLMs)
as programming assistants to generate data analysis scripts, execute code refactoring, and support minor implementation tasks.
In all instances, the generated source code was strictly verified by a human to ensure correctness. 
Additionally, we utilized LLMs as writing assistants to optimize vocabulary choices, refine grammatical phrasing, and assist in structural LaTeX formatting across the manuscript without altering any underlying technical content.

\end{document}